\documentclass{article}

\usepackage{CJKutf8} 
\usepackage[utf8]{inputenc}

\usepackage[final,nonatbib]{neurips_2022}

\usepackage[utf8]{inputenc} 
\usepackage[T1]{fontenc}    
\usepackage{hyperref}       
\usepackage{url}            
\usepackage{booktabs}       
\usepackage{amsfonts}       
\usepackage{nicefrac}       
\usepackage{microtype}      
\usepackage{xcolor}         

\usepackage{wrapfig}
\usepackage{times}
\usepackage{latexsym}

\definecolor{darkblue}{rgb}{0.0,0.0,0.5}

\usepackage{amsmath}
\usepackage{array}
\newcolumntype{L}{>{$}l<{$}}
\newcolumntype{C}{>{$}c<{$}}
\newcolumntype{R}{>{$}r<{$}}

\usepackage{multirow}
\usepackage{tabularx, caption, boldline}
\usepackage{graphicx}
\usepackage{cellspace}
\usepackage{algpseudocode}  
\usepackage{multicol}  
\usepackage{flexisym}

\usepackage{microtype}

\usepackage{dsfont}

\makeatletter
\def\hlinewd#1{%
\noalign{\ifnum0=`}\fi\hrule \@height #1 %
\futurelet\reserved@a\@xhline}
\makeatother

\usepackage{times}
\usepackage{latexsym}

\usepackage{algpseudocode}  
\usepackage{amsmath}
\usepackage{multicol}  
\usepackage{multirow} 
\usepackage{graphicx}  
\usepackage{array}
\usepackage{arydshln}
\usepackage{booktabs}
\usepackage{textcomp}

\usepackage{amssymb}
\usepackage{pifont}

\usepackage{cleveref}
\crefformat{section}{\S#2#1#3} 
\crefformat{subsection}{\S#2#1#3}
\crefformat{subsubsection}{\S#2#1#3}

\usepackage{microtype}

\usepackage{adjustbox}
\usepackage{multicol}  
\usepackage{multirow} 
\usepackage{amsmath}
\usepackage{amsfonts}
\usepackage{makecell}
\usepackage{algpseudocode} 
\usepackage{verbatim}
\usepackage{mathtools}
\DeclareMathOperator*{\argmax}{arg\,max}

\usepackage{booktabs}

\usepackage{makecell}
\usepackage{cleveref}
\usepackage[normalem]{ulem}

\usepackage{listings}

\definecolor{Code}{rgb}{0,0,0} 
\definecolor{Decorators}{rgb}{0.5,0.5,0.5} 
\definecolor{Numbers}{rgb}{0.5,0,0} 
\definecolor{MatchingBrackets}{rgb}{0.25,0.5,0.5} 
\definecolor{Keywords}{rgb}{0,0,1} 
\definecolor{self}{rgb}{0,0,0} 
\definecolor{Strings}{rgb}{0,0.63,0} 
\definecolor{Comments}{rgb}{0.63,0,0} 
\definecolor{Backquotes}{rgb}{0,0,0} 
\definecolor{Classname}{rgb}{0,0,0} 
\definecolor{FunctionName}{rgb}{0,0,0} 
\definecolor{Operators}{rgb}{0,0,0} 
\definecolor{Background}{rgb}{0.99,0.99,0.99} 

\usepackage{setspace}
\usepackage{CJKutf8}

\usepackage{times}
\usepackage{latexsym}

\usepackage{footnote}
\makesavenoteenv{tabular}
\makesavenoteenv{table}

\usepackage[hang,flushmargin]{footmisc}

\usepackage[linesnumbered,ruled]{algorithm2e}
\SetAlFnt{\small}
\SetAlCapNameFnt{\normalsize}
\makeatletter
\newcommand{\nosemic}{\renewcommand{\@endalgocfline}{\relax}}
\newcommand{\dosemic}{\renewcommand{\@endalgocfline}{\algocf@endline}}
\let\oldnl\nl
\newcommand{\nonl}{\renewcommand{\nl}{\let\nl\oldnl}}
\makeatother

\usepackage{enumitem}
\usepackage{tablefootnote}

\usepackage{tabularx}
\makeatletter
\def\hlinewd#1{%
\noalign{\ifnum0=`}\fi\hrule \@height #1 %
\futurelet\reserved@a\@xhline}
\makeatother

\usepackage{minitoc}

\usepackage{cleveref}
\crefformat{section}{\S#2#1#3}
\crefformat{subsection}{\S#2#1#3}
\crefformat{subsubsection}{\S#2#1#3}
\crefrangeformat{section}{\S\S#3#1#4 to~#5#2#6}
\crefmultiformat{section}{\S\S#2#1#3}{ and~#2#1#3}{, #2#1#3}{ and~#2#1#3}
\usepackage{refstyle}
\Crefformat{figure}{#2Figure~#1#3}
\Crefmultiformat{figure}{Figures.~#2#1#3}{ and~#2#1#3}{, #2#1#3}{ and~#2#1#3}
\Crefformat{table}{#2Table~#1#3}
\Crefmultiformat{table}{Tables~#2#1#3}{ and~#2#1#3}{, #2#1#3}{ and~#2#1#3}
\Crefformat{appendix}{Appx.~\S#2#1#3}
\crefformat{algorithm}{Alg.~#2#1#3}

\definecolor{NavyBlue}{rgb}{0.1, 0.4, 0.8}
\hypersetup{colorlinks = true, linkcolor = NavyBlue,
            urlcolor  = gray,
            citecolor = NavyBlue,
            anchorcolor = NavyBlue}

\title{A Contrastive Framework for Neural Text Generation}

\author{
 \textbf{Yixuan Su}$^\spadesuit$  \quad
 \textbf{Tian Lan}$^\diamondsuit$  \quad
 \textbf{Yan Wang}$^\diamondsuit$  \quad
 \textbf{Dani Yogatama}$^\clubsuit$  \quad
 \\
 \textbf{Lingpeng Kong}$^\heartsuit$ \quad
 \textbf{Nigel Collier}$^\spadesuit$\\
 $^\spadesuit$Language Technology Lab, University of Cambridge\\
 $^\diamondsuit$Tencent AI Lab \ \ \ \ \ $^\clubsuit$DeepMind\\
 $^\heartsuit$Department of Computer Science, The University of Hong Kong\\
 {\tt \{ys484,nhc30\}@cam.ac.uk}\\
 {\tt lantiangmftby@gmail.com, 
 yanwang.branden@gmail.com}\\
 {\tt dyogatama@deepmind.com, lpk@cs.hku.hk}
}

\begin{document}
\maketitle

\doparttoc 
\faketableofcontents

\begin{abstract}
Text generation is of great importance to many natural language processing applications. However, maximization-based decoding methods (e.g., beam search) of neural language models often lead to degenerate solutions---the generated text is unnatural and contains undesirable repetitions. 
Existing approaches introduce stochasticity via sampling or modify training objectives to decrease the probabilities of certain tokens (e.g., unlikelihood training).
However, they often lead to solutions that lack coherence. 
In this work, we show that an underlying reason for model degeneration is the anisotropic distribution of token representations. We present a contrastive solution: (i) \textit{SimCTG}, a contrastive training objective to calibrate the model's representation space, and (ii) a decoding method---\textit{contrastive search}---to encourage diversity while maintaining coherence in the generated text. Extensive experiments and analyses on three benchmarks from two languages demonstrate that our proposed approach significantly outperforms current state-of-the-art text generation methods as evaluated by both human and automatic metrics.\footnote{Our code and models are publicly available at \url{https://github.com/yxuansu/SimCTG}.}
\end{abstract}

\section{Introduction}
\label{sec:introduction}
Open-ended neural text generation with Transformer \cite{DBLP:conf/nips/VaswaniSPUJGKP17} is an indispensable component in various natural language applications, such as story generation \cite{DBLP:conf/acl/LewisDF18,su2022language}, contextual text completion \cite{radford2019language}, and dialogue systems \cite{DBLP:journals/taslp/SuWCBKC21}. However, the conventional approach of training a language model with maximum likelihood estimation (MLE) and decoding the most likely sequence is often not sufficient \cite{DBLP:conf/iclr/HoltzmanBDFC20,DBLP:conf/iclr/WelleckKRDCW20}. Specifically, this modelling formulation often leads to the problem of \textit{degeneration}, i.e., the generated texts from the language model tend to be dull and contain undesirable repetitions at different levels (e.g., token-, phrase-, and sentence-level) \cite{DBLP:journals/corr/abs-1902-00098}. To alleviate this problem, previous solutions modify the decoding strategy by sampling from \textit{less} likely vocabularies \cite{DBLP:conf/acl/LewisDF18,DBLP:conf/iclr/HoltzmanBDFC20}. While reducing the generated repetition, these sampling methods introduce another critical  problem (\textit{semantic inconsistency})---the sampled text tends to diverge from or even contradict to the original semantics defined by the human-written prefix \cite{DBLP:conf/iclr/BasuRKV21}. Another approach addresses the degeneration problem by modifying the model's output vocabulary distribution with unlikelihood training \cite{DBLP:conf/iclr/WelleckKRDCW20}.

\begin{figure*}[h] 
  \centering    
  \setlength{\abovecaptionskip}{3pt}
  \includegraphics[width=0.95\textwidth]{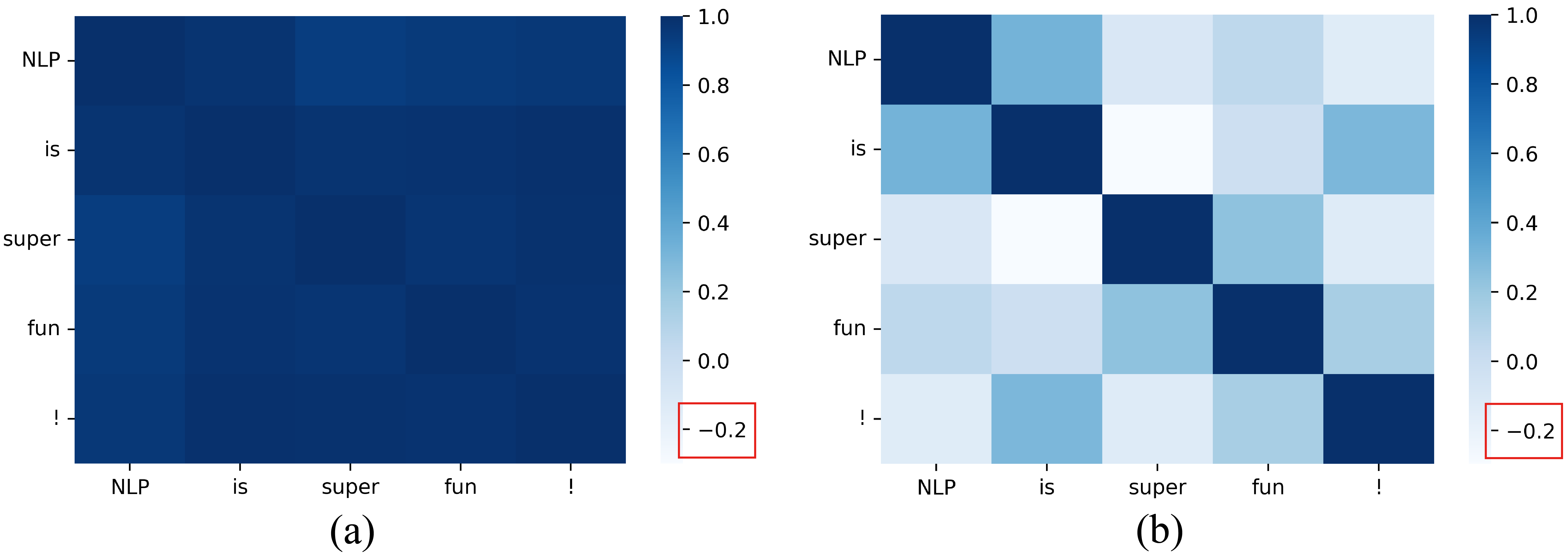}
  \caption{Token cosine similarity matrix of (a) GPT-2 and (b) SimCTG. (best viewed in color)}
  \label{fig:heatmap_example}
  \vspace{-1.5mm}
\end{figure*}

In this work, we argue that the degeneration of neural language models stems from  the \textit{anisotropic} distribution of token representations, i.e., their representations reside in a narrow subset of the entire space \cite{DBLP:conf/emnlp/Ethayarajh19,DBLP:conf/icml/DongCL21,DBLP:journals/corr/abs-2111-04198}. In Figure \ref{fig:heatmap_example}(a), we showcase a cosine similarity matrix of token representations (taken from the output layer of the Transformer) produced by GPT-2. We see that the cosine similarities between tokens within a sentence are over \textbf{0.95}, meaning that these representations are close to each other. Such high similarity is undesirable as it can naturally cause the model to generate repetitive tokens at different steps. 
In an ideal setting, the token representations should follow an isotropic distribution, i.e., the token similarity matrix should be sparse and the representations of distinct tokens should be discriminative as shown in Figure \ref{fig:heatmap_example}(b). Moreover, during the decoding process, the sparseness of the token similarity matrix of the generated text should be preserved to avoid model degeneration.

Based on the above motivations, we present \textit{SimCTG} (a \underline{\textbf{sim}}ple \underline{\textbf{c}}ontrastive framework for neural \underline{\textbf{t}}ext \underline{\textbf{g}}eneration) that encourages the model to learn discriminative and isotropic token representations. 
We also present a novel decoding strategy to complement SimCTG, \textit{contrastive search}.
The key intuitions behind contrastive search are: (i) at each decoding step, the output should be selected from the set of most probable candidates predicted by the model to better maintain the semantic coherence between the generated text and the human-written prefix, and (ii) the sparseness of the token similarity matrix of the generated text should be preserved to avoid degeneration.


We conduct comprehensive experiments on three widely used benchmarks. We show that our approach is generalizable to different tasks and different languages (\cref{sec:experiment} and \cref{sec:dialogue_experiment}) as well as different model sizes (\cref{sec:human_evaluation_detail} and Appendix~\ref{appendix:different_model_experiment}). Specifically, the experimental results verify that SimCTG improves the intrinsic qualities of the language model, as evaluated by perplexity and token prediction accuracy (\cref{sec:automatic_evaluation_result} and Appendix~\ref{appendix:different_model_experiment}). Moreover, we demonstrate that the proposed contrastive search significantly outperforms previous state-of-the-art decoding methods in both human and automatic evaluations (\cref{sec:experiment} and \cref{sec:dialogue_experiment}). Furthermore, we provide in-depth analyses to get better insights on the inner-workings of our proposed approach (\cref{sec:detailed_analysis_section}).

\section{Background}
\subsection{Language Modelling}
The goal of language modelling is to learn a probability distribution $p_{\theta}(\boldsymbol{x})$ over a variable-length text sequence $\boldsymbol{x}=\{x_1, ..., x_{|\boldsymbol{x}|}\}$, where $\theta$ denotes model parameters. Typically, the maximum likelihood estimation (MLE) objective is used to train the language model which is defined as
\begin{equation}
    \label{eq:mle}
    \mathcal{L}_{\textup{MLE}} = -\frac{1}{|\boldsymbol{x}|}\sum_{i=1}^{|\boldsymbol{x}|}\log p_{\theta}(x_i|\boldsymbol{x}_{<i}).
\end{equation}

However, as observed in many recent studies \cite{DBLP:conf/emnlp/Ethayarajh19,DBLP:conf/icml/DongCL21,DBLP:journals/corr/abs-2111-04198}, training with likelihood maximization objective often yields an anisotropic distribution of model representations (especially for Transformer-based models) that undermines the model's capacity.

\subsection{Open-ended Text Generation}
In this work, we focus on studying the task of open-ended text generation due to its generality in various applications, such as story generation \cite{DBLP:conf/acl/LewisDF18,su2022language}, contextual text completion \cite{radford2019language}, poetry generation \cite{DBLP:conf/acl/LiZLS20}, and dialogue systems \cite{DBLP:journals/taslp/SuWCBKC21}. Formally, conditioned on a human-written prefix (i.e., context) $\boldsymbol{x}$, the task is to decode a continuation $\hat{\boldsymbol{x}}$ from the language model and the resulting text is $\{x_1,..,x_{|\boldsymbol{x}|},\hat{x}_{|\boldsymbol{x}|+1},...,\hat{x}_{|\boldsymbol{x}|+|\hat{\boldsymbol{x}}|}\}$. Typically, there are two classes of methods used for decoding, which are (1) deterministic methods and (2) stochastic methods.

\textbf{Deteriminstic Methods.} Two widely used deterministic approaches are greedy and beam search which aim to select the text continuation with highest probability based on the model's probability distribution $p_{\theta}$. However, solely maximizing the output probability often leads to dullness \cite{DBLP:conf/naacl/LiGBGD16} and degeneration \cite{DBLP:conf/acl/LewisDF18,DBLP:conf/iclr/HoltzmanBDFC20} in the generated text. 

\textbf{Stochastic Methods.} To remedy the issues of deterministic decoding, several approaches have been proposed to sample from $p_{\theta}$. To avoid sampling from the unreliable tail of distribution, Fan \emph{et al.}~\cite{DBLP:conf/acl/LewisDF18} proposed top-$k$ sampling which draws sample from the vocabulary subset $V^{(k)}$ that maximizes $\sum_{v\in V^{(k)}}p_{\theta}(v|\boldsymbol{x})$. Here, $|V^{(k)}|=k$ and $\boldsymbol{x}$ is the prefix context. Differently, the current state-of-the-art nucleus sampling \cite{DBLP:conf/iclr/HoltzmanBDFC20} draws sample from the smallest vocabulary subset $U$ with total probability mass above a threshold $p\in[0,1]$; i.e., $U$ is the smallest vocabulary subset such that $\sum_{v\in U}p_{\theta}(v|\boldsymbol{x})\geq p$. While the sampling approaches help to alleviate model degeneration, the intrinsic stochasticity in these methods could cause the semantic meaning of the sampled text to diverge from or even contradict to the human-written prefix \cite{DBLP:conf/iclr/BasuRKV21}.

\section{Methodology}
In this section, we first present how to apply contrastive learning to calibrate the representation space of the language model. Then, we introduce our proposed contrastive search decoding algorithm. 

\subsection{Contrastive Training}
\label{sec:contrastive_training}
Our goal is to encourage the language model to learn discriminative and isotropic token representations. To this end, we introduce a contrastive objective $\mathcal{L}_{\textup{CL}}$ into the training of the language model. Specifically, given a variable-length sequence $\boldsymbol{x}=\{x_1, ..., x_{|\boldsymbol{x}|}\}$, the $\mathcal{L}_{\textup{CL}}$ is defined as 
\begin{equation}
    \label{eq:cl}
    \mathcal{L}_{\textup{CL}} = \frac{1}{|\boldsymbol{x}|\times(|\boldsymbol{x}| - 1)}\sum_{i=1}^{|\boldsymbol{x}|}\sum_{j=1,j\neq i}^{|\boldsymbol{x}|}\max\{0,\rho - s(h_{x_i}, h_{x_i}) + s(h_{x_i}, h_{x_j})\},
\end{equation}
where $\rho\in[-1,1]$ is a pre-defined margin  and $h_{x_i}$ is the representation of token $x_i$ produced by the model. The similarity function $s$ computes the cosine similarity between token representations as
\begin{equation}
    \label{eq:cosine}
    s(h_{x_i}, h_{x_j}) = \frac{h_{x_i}^\top h_{x_j}}{\|h_{x_i}\|\cdot\|h_{x_j}\|}.
\end{equation}
Intuitively, by training with $\mathcal{L}_{\textup{CL}}$, the model learns to pull away the  distances between representations of distinct tokens.\footnote{By definition, the cosine similarity $s(h_{x_i}, h_{x_i})$ of the identical token $x_i$ is $1.0$.} Therefore, a discriminative and isotropic model representation space can be obtained. The overall training objective $\mathcal{L}_{\textup{SimCTG}}$ is then defined as
\begin{equation}
    \label{eq:simctg}
    \mathcal{L}_{\textup{SimCTG}} = \mathcal{L}_{\textup{MLE}} + \mathcal{L}_{\textup{CL}},
\end{equation}
where the maximum likelihood estimation (MLE) objective $\mathcal{L}_{\textup{MLE}}$ is described in Eq. (\ref{eq:mle}). Note that, when the margin $\rho$ in $\mathcal{L}_{\textup{CL}}$ equals to $0$, the $\mathcal{L}_{\textup{SimCTG}}$ degenerates to the vanilla MLE objective $\mathcal{L}_{\textup{MLE}}$. 

\subsection{Contrastive Search}
We propose a novel decoding method, \textit{contrastive search}. At each decoding step, the key ideas of contrastive search are (i) the generated output should be selected from the set of most probable candidates predicted by the model; and (ii) the generated output should be discriminative enough with respect to the previous context. In this way, the generated text can (i) better maintain the semantic coherence with respect to the prefix while (ii) avoiding model degeneration.

Formally, given the previous context $\boldsymbol{x}_{<t}$, at time step $t$, the selection of the output $x_t$ follows
\begin{equation}
    \label{eq:score}
    x_t = \argmax_{v\in V^{(k)}}\bigg\{(1 - \alpha)\times \underbrace{p_{\theta}(v|\boldsymbol{x}_{<t})}_{\textup{model confidence}} -  \: \alpha \times \underbrace{(\max\{s(h_v, h_{x_j}):1\leq j \leq t-1\})}_{\textup{degeneration penalty}}\bigg\},
\end{equation}
where $V^{(k)}$ is the set of top-$k$ predictions from the model's probability distribution $p_{\theta}(\cdot|\boldsymbol{x}_{<t})$ 
and $k$ is typically set as 3$\sim$10. In Eq. (\ref{eq:score}), the first term, \textit{model confidence}, is the probability of candidate $v$ predicted by the model. The second term, \textit{degeneration penalty}, measures how discriminative of candidate $v$ with respect to the previous context $\boldsymbol{x}_{<t}$ and $s$ is defined in Eq. (\ref{eq:cosine}). Specifically, it is defined as the maximum cosine similarity between the representation of $v$ and that of all tokens in $\boldsymbol{x}_{<t}$. Here, the candidate representation $h_v$ is computed by the model given the concatenation of $\boldsymbol{x}_{<t}$ and $v$. Intuitively, a larger degeneration penalty of $v$ means it is more similar to the context, therefore more likely leading to model degeneration. The hyperparameter $\alpha\in[0,1]$ regulates the importance of these two components. When $\alpha=0$, contrastive search degenerates to the greedy search method.


\section{Document Generation}
\label{sec:experiment}
We first evaluate our approach on the task of open-ended document generation.

\textbf{Model and Baselines.} Our proposed approach is architecture-agnostic and can be applied to any generation model. In this work, we evaluate our method on the representative GPT-2 model \cite{radford2019language}. Specifically, we fine-tune GPT-2 on the evaluated benchmark (detailed below) with the proposed objective $\mathcal{L}_{\textup{SimCTG}}$ (Eq. (\ref{eq:simctg})) and generate the text continuation with different decoding methods. We perform experiments using the base model (117M parameters) which consists of 12 Transformer layers \cite{DBLP:conf/nips/VaswaniSPUJGKP17} with 12 attention heads.\footnote{In Appendix~\ref{appendix:different_model_experiment}, we demonstrate the experimental results of our approach on other language models.} We compare our approach with two strong baselines: (1) GPT-2 fine-tuned with the standard MLE objective (Eq. (\ref{eq:mle})); and (2) GPT-2 fine-tuned with unlikelihood objective \cite{DBLP:conf/iclr/WelleckKRDCW20}.\footnote{The unlikelihood baseline is implemented with the official code, which can be found at \url{https://github.com/facebookresearch/unlikelihood_training}.} Our implementation is based on the Huggingface Library \cite{DBLP:journals/corr/abs-1910-03771}.

\textbf{Evaluation Benchmark.} We conduct experiments on the   Wikitext-103 dataset \cite{DBLP:conf/iclr/MerityX0S17} which contains a large collection of Wikipedia articles with over 100 million words and 260 thousands unique tokens. Wikitext-103 is a document-level dataset and has been widely used for the evaluation of large-scale language modelling \cite{DBLP:conf/acl/DaiYYCLS19,DBLP:conf/iclr/KhandelwalLJZL20,DBLP:journals/tacl/YogatamadK21}.

\textbf{Training.} For our SimCTG and the MLE baseline, we fine-tune the models on Wikitext-103 for 40k training steps. For the unlikelihood baseline, following Welleck \emph{et al.}~\cite{DBLP:conf/iclr/WelleckKRDCW20}, we first fine-tune the model with the token-level unlikelihood objective for 38.5k steps and then with the sequence-level unlikelihood objective for 1.5k steps. Therefore, the overall training steps of all compared methods are the same. The batch size is set as 128 and the training samples are truncated to a maximum length of 256. We optimize the model with Adam optimizer \cite{DBLP:journals/corr/KingmaB14} and a learning rate of 2e-5.

\textbf{Decoding.} We evaluate the models by producing text continuations given the prefixes from the test set. In the experiments, the lengths of the prefix and the generated continuation are set as 32 and 128, respectively. We test different models with various decoding methods. For deterministic method, we use greedy search and beam search with a beam size of $10$. For stochastic method, we use the current state-of-the-art nucleus sampling \cite{DBLP:conf/iclr/HoltzmanBDFC20} with $p=0.95$. For the proposed contrastive search, the $k$ and $\alpha$ in Eq. (\ref{eq:score}) are set as $8$ and $0.6$.\footnote{In Appendix \ref{appendix:ablation_study}, we provide detailed ablation studies on the effect of both $k$ and $\alpha$ in contrastive search.}  
The hyperparameters of different methods are selected based on their optimal MAUVE (detailed in \cref{sec:generation_quality_metric}) performance on the validation set.

\subsection{Evaluation Metrics}
We perform evaluation from two aspects: (1) \textit{language modelling quality} which measures the intrinsic quality of the model; and (2) \textit{generation quality} which measures the quality of the generated text. 

\subsubsection{Language Modelling Quality}
Following Welleck \emph{et al.}~\cite{DBLP:conf/iclr/WelleckKRDCW20}, we report the results of the model on the metrics below.

\textbf{Perplexity.} The model perplexity (\textbf{ppl}) on the test set of Wikitext-103.

\textbf{Prediction Accuracy.} It is defined as: $\textup{\textbf{acc}}=\frac{1}{\sum_{\boldsymbol{x}\in\mathcal{D}}|\boldsymbol{x}|}\sum_{\boldsymbol{x}\in\mathcal{D}}\sum_{t=1}^{|\boldsymbol{x}|}\mathds{1}[\argmax p_{\theta}(x|\boldsymbol{x}_{<t})=x_t]$, where $\mathcal{D}$ is the Wikitext-103 test set, $\boldsymbol{x}_{<t}$ is the prefix, and $x_t$ is the reference token at time step $t$.

\textbf{Prediction Repetition.} The fraction of next-token (top-1) predictions that occur in the prefix which is defined as: $\textup{\textbf{rep}} = \frac{1}{\sum_{\boldsymbol{x}\in\mathcal{D}}|\boldsymbol{x}|}\sum_{\boldsymbol{x}\in\mathcal{D}}\sum_{t=1}^{|\boldsymbol{x}|}\mathds{1}[\argmax p_{\theta}(x|\boldsymbol{x}_{<t})\in \boldsymbol{x}_{<t}]$. 

In addition, the next token repetitions that do not equal to the ground truth token: $\textup{\textbf{wrep}} = \frac{1}{\sum_{\boldsymbol{x}\in\mathcal{D}}|\boldsymbol{x}|}\sum_{\boldsymbol{x}\in\mathcal{D}}\sum_{t=1}^{|\boldsymbol{x}|}\mathds{1}[\argmax p_{\theta}(x|\boldsymbol{x}_{<t}) \in \boldsymbol{x}_{<t} \: \wedge \neq x_t]$ is also reported.

\subsubsection{Generation Quality}
\label{sec:generation_quality_metric}
\textbf{Generation Repetition.} This metric measures the sequence-level repetition as the portion of duplicate $n$-grams in the generated text \cite{DBLP:conf/iclr/WelleckKRDCW20}. For a generated text continuation $\hat{\boldsymbol{x}}$, the repetion at $n$-gram level is defined as: $\textup{\textbf{rep-n}}=100 \times (1.0 - \frac{|\textup{unique n-grams}(\hat{\boldsymbol{x}})|}{|\textup{total n-grams}(\hat{\boldsymbol{x}})|})$. 

\textbf{Diversity.} This metric takes into account the generation repetition at different $n$-gram levels and it is defined as: $\textup{\textbf{diversity}}=\prod_{n=2}^{4}(1.0-\frac{\textup{rep-n}}{100})$. It can be deemed as an overall assessment of model degeneration. A lower diversity means a more severe degeneration of the model.

\textbf{MAUVE} \cite{pillutla2021mauve} is a metric that measures the token distribution closeness between the generated text and human-written text. A higher MAUVE score means the model generates more human-like texts.

\textbf{Semantic Coherence.} To automatically measure the semantic coherence (i.e., consistency) between the prefix and the generated text, we employ the advanced sentence embedding method, SimCSE \cite{DBLP:conf/emnlp/GaoYC21}. Specifically, given the prefix $\boldsymbol{x}$ and the generated text $\hat{\boldsymbol{x}}$, the coherence score is defined as:  
$\textup{\textbf{coherence}}=v_{\boldsymbol{x}}^\top v_{\hat{\boldsymbol{x}}}/(\|v_{\boldsymbol{x}}\|\cdot\|v_{\hat{\boldsymbol{x}}}\|)$, where $v_{\boldsymbol{x}}=\textup{SimCSE}(\boldsymbol{x})$ and $v_{\hat{\boldsymbol{x}}}=\textup{SimCSE}(\hat{\boldsymbol{x}})$.

\textbf{Perplexity of Generated Text.} Lastly, we evaluate the perplexity of the generated text $\hat{\boldsymbol{x}}$ given the prefix $\boldsymbol{x}$, which is defined as: $\textup{\textbf{gen-ppl}}=2^{f(\mathcal{D},\theta)}$ and $f(\mathcal{D},\theta)=\frac{1}{\sum_{\boldsymbol{x}\in\mathcal{D}}|\hat{\boldsymbol{x}}|}\sum_{\boldsymbol{x}\in\mathcal{D}}\log_{2}p_{\theta}(\hat{\boldsymbol{x}}|\boldsymbol{x})$. Importantly, the optimal approach should produce text which has a perplexity \textit{close} to that of the human-written text \cite{DBLP:conf/iclr/HoltzmanBDFC20}.  
A high gen-ppl means the generated text is very \textit{unlikely} given the prefix, therefore being low quality. In contrastive, a low gen-ppl means the generated text has a low diversity and gets stuck in repetitive loops \cite{DBLP:conf/iclr/HoltzmanBDFC20}.
We use the model $\theta$ trained with $\mathcal{L}_{\textup{SimCTG}}$ to measure the gen-ppl of different approaches, therefore making sure the numbers are comparable with each other.\footnote{We obtain similar gen-ppl results and can draw the same conclusion when using the model trained with MLE and Unlikelihood. Therefore, we only include the results acquired by the model trained with $\mathcal{L}_{\textup{SimCTG}}$ in Table \ref{tb:main_result}. We refer to Appendix \ref{appendix:gen_ppl} for the gen-ppl results obtained by the MLE and Unlikelihood models.}

\begin{table*}[t]
    \small
	\centering  
	\renewcommand{\arraystretch}{1.2}
	\setlength{\tabcolsep}{6pt}
	\scalebox{0.75}{
	\begin{tabular}{ccccccccccccc}
		\hlinewd{0.75pt}
		\multirow{2}{*}{\textbf{Model}}&\multicolumn{4}{c}{Language Modelling Quality}&&\multicolumn{7}{c}{Generation Quality}\\
		\cmidrule(lr){2-5}
		\cmidrule(lr){6-13}
		&ppl$\downarrow$&acc$\uparrow$&rep$\downarrow$&wrep$\downarrow$& Method&rep-2$\downarrow$&rep-3$\downarrow$&rep-4$\downarrow$&diversity$\uparrow$&MAUVE$\uparrow$&coherence$\uparrow$&gen-ppl\\
		\hline
		\multirow{4}{*}{MLE}&\multirow{4}{*}{24.32}&\multirow{4}{*}{39.63}&\multirow{4}{*}{52.82}&\multirow{4}{*}{29.97}&greedy&69.21&65.18&62.05&0.04&0.03&0.587&7.32\\
		&&&&&beam&71.94&68.97&66.62&0.03&0.03&0.585&6.42\\
		&&&&&nucleus&4.45&0.81&0.43&0.94&0.90&0.577&49.71\\
		&&&&&contrastive&44.20&37.07&32.44&0.24&0.18&0.599&9.90\\
        \hline
        \multirow{4}{*}{Unlike.}&\multirow{4}{*}{28.57}&\multirow{4}{*}{38.41}&\multirow{4}{*}{\textbf{51.23}}&\multirow{4}{*}{\textbf{28.57}}&greedy&24.12&13.35&8.04&0.61&0.69&0.568&37.82\\
		&&&&&beam&11.83&5.11&2.86&0.81&0.75&0.524&34.73\\
		&&&&&nucleus&4.01&0.80&0.42&\textbf{0.95}&0.87&0.563&72.03\\
		&&&&&contrastive&7.48&3.23&1.40&0.88&0.83&0.574&43.61\\
        \hline
		\multirow{4}{*}{SimCTG}&\multirow{4}{*}{\textbf{23.82}}&\multirow{4}{*}{\textbf{40.91}}&\multirow{4}{*}{51.66}&\multirow{4}{*}{28.65}&greedy &67.36&63.33&60.17&0.05&0.05&0.596&7.16\\
		&&&&&beam&70.32&67.17&64.64&0.04&0.06&0.591&6.36\\
		&&&&&nucleus&4.05&0.79&0.37&0.94&0.92&0.584&47.19\\
		&&&&&contrastive&\textbf{3.93}&\textbf{0.78}&\textbf{0.31}&\textbf{0.95}&\textbf{0.94}&\textbf{0.610}&\textbf{18.26}\\
		\hline
		Human&-&-&36.19&-&-&3.92&0.88&0.28&0.95&1.00&0.644&24.01\\
		\hlinewd{0.75pt}
	\end{tabular}}
    \caption{Evaluation results on Wikitext-103 test set. ``Unlike.'' denotes the model trained with unlikelihood objective. $\uparrow$ means higher is better and $\downarrow$ means lower is better.}
    	\vspace{-1.5mm}
	\label{tb:main_result}
\end{table*}

\subsection{Results}
\label{sec:automatic_evaluation_result}
The experimental results on Wikitext-103 are shown in Table \ref{tb:main_result}. 

\textbf{Language Modelling Quality.} From the results, we observe that SimCTG achieves the best perplexity and next token accuracy. The reason is that, with more discriminative representations, SimCTG is less confusing when making next token predictions, leading to the improved model performance. 
On the rep and wrep metrics, the unlikelihood model yields the best result but at the expense of unfavorable performance drops in the perplexity and next token accuracy. 

\textbf{Generation Quality.} Firstly, on the rep-n and diversity metrics, SimCTG + contrastive search obtains the best result, suggesting it best addresses the degeneration problem. Secondly, the MAUVE score demonstrates that SimCTG + contrastive search generates texts that are closest to human-written texts in terms of token distribution. Thirdly, among all methods, SimCTG + contrastive search is the only approach that achieves over 0.6 coherence score, showing it produces semantically consistent text with respect to the prefix. Lastly, the gen-ppl metric also validates the superiority of SimCTG + contrastive search as it obtains notably better generation perplexity comparing with other approaches.

Moreover, from the results of MLE and Unlikelihood baselines, we see that contrastive search still brings performance boost as compared with greedy and beam search. However, the performance gain still lags behind SimCTG, which demonstrates the necessity of contrastive training.  The underlying reason is that, without using the contrastive objective $\mathcal{L}_{\textup{CL}}$ (Eq. (\ref{eq:cl})), the token representations obtained by MLE or Unlikelihood are less discriminative (\cref{sec:self_similarity}). Therefore, the degeneration penalty (Eq. (\ref{eq:score})) of different candidates are less distinguishable and the selection of output is dominated by the model confidence, making contrastive search less effective.

\begin{table*}[h]
    \small
	\centering  
	\renewcommand{\arraystretch}{1.2}
	\setlength{\tabcolsep}{6pt}
	\scalebox{1.0}{
	\begin{tabular}{ccccc}
		\hlinewd{0.75pt}
        \textbf{Model}&Decoding Method&Coherence&Fluency&Informativeness\\
        \hline
        Agreement&-&0.51&0.64&0.70\\
        \hline
        \multirow{2}{*}{MLE}&nucleus&2.92&3.32&3.91\\
        &contrastive&2.78&2.29&2.56\\
        \hline
        \multirow{2}{*}{Unlikelihood}&nucleus&2.59&3.02&3.58\\
        &contrastive&2.76&2.90&3.35\\
        \hline
        \multirow{2}{*}{SimCTG}&nucleus&2.96&3.34&3.96\\
        &contrastive&3.25$^{\bigstar}$&3.57$^{\bigstar}$&3.96\\
        \hline
        \multirow{2}{*}{SimCTG-large}&nucleus&3.01&3.37&\textbf{3.98}\\
        &contrastive&\textbf{3.33}$^{\bigstar}$&\textbf{3.66}$^{\bigstar}$&\textbf{3.98}\\
        \hline
        Human&-&3.70&3.71&4.21\\
		\hlinewd{0.75pt}
	\end{tabular}}
    \caption{Human evaluation results. ${\bigstar}$ results significantly outperforms the results of nucleus sampling with different models (Sign Test with p-value < 0.05).}
    	\vspace{-1.5mm}
	\label{tb:human_evaluation}
\end{table*}

\subsection{Human Evaluation}
\label{sec:human_evaluation_detail}
We also conduct a human evaluation with the help of graders proficient in English from a third-party grading platform. We randomly select 200 prefixes with length of 32 from the test set of Wikitext-103. For each prefix, we use different models (MLE, Unlikelihood, and SimCTG) with two decoding methods (nucleus sampling and contrastive search) to generate text continuations with length of 128. To examine the generality of our approach across different model sizes, we include a large size SimCTG (i.e., \textbf{SimCTG-large}) which is obtained by fine-tuning the GPT-2-large model that consists of 36 Transformer layers with 20 attention heads. All generated results, plus the reference text, are randomly shuffled and evaluated by five graders, which results in 9,000 annotated samples in total. The evaluation follows a 5-point Likert scale (1, 2, 3, 4, or 5) for each of the following features:\footnote{We refer to Appendix \ref{sec:evaluation_guideline} for more  details of human evaluation.}
\begin{itemize}[noitemsep,topsep=1pt]
    \itemsep 0em 
    \item \textbf{Coherence}: Whether the generated text is semantically consistent with the prefix.
    \item \textbf{Fluency}: Whether the generated text is fluent and easy to understand.
    \item \textbf{Informativeness}: Whether the generated text is diverse and contains interesting content. 
\end{itemize}

Table \ref{tb:human_evaluation} presents the human evaluation results, with the first row showing strong inter-annotator agreements as measured by Fleiss$\textprime$ kappa coefficient \cite{fleiss1971mns}. Firstly, we see that, directly applying contrastive search with MLE or Unlikelihood model does not yield satisfactory results. This is due to the anisotropic nature of their representation space as discussed in Section \cref{sec:automatic_evaluation_result}. Secondly, the coherence score of Unlikelihood model is notably lower than MLE and SimCTG, suggesting it generates the most \textit{unlikely} results which is also shown by its generation perplexity (gen-ppl) in Table \ref{tb:main_result}. Furthermore, the results of SimCTG + contrastive search significantly outperforms nucleus sampling with different models  in terms of coherence and fluency (Sign Test with p-value < 0.05). Lastly, SimCTG-large + contrastive search achieves the best performance across the board and even performs comparably with human-written text on the fluency metric (Sign Test with p-value > 0.4). This reveals the clear generalization ability of our approach to large size models and future work could focus on extending it to models that contain over billions of parameters such as GPT-3 \cite{DBLP:conf/nips/BrownMRSKDNSSAA20}.

\section{Open-domain Dialogue Generation}
\label{sec:dialogue_experiment}
To test the generality of our approach across different tasks and languages, we then evaluate our method on the task of open-domain dialogue generation. In this task, given a multi-turn dialogue context (where each turn is an user utterance), the model is asked to generate an adequate response that is semantically consistent with the context. Here, the dialogue context is deemed as the prefix.

\textbf{Benchmark and Baselines.} We conduct experiments on two benchmark datasets from two languages (i.e., Chinese and English). For the Chinese benchmark, we use the LCCC dataset \cite{DBLP:conf/nlpcc/WangKZHJZH20}. For the English Benchmark, we use the DailyDialog dataset \cite{DBLP:conf/ijcnlp/LiSSLCN17}.

We compare the GPT-2 models fine-tuned with SimCTG and MLE.\footnote{We acknowledge that there are other GPT-like models (e.g., Zhang \emph{et al.}~\cite{DBLP:conf/acl/ZhangSGCBGGLD20} and Thoppilan \emph{et al.}~\cite{DBLP:journals/corr/abs-2201-08239}) that are designed for dialogue generation. We leave the test of our approach on these models to our future work.} Specifically, for the Chinese benchmark (i.e., LCCC), we use a publicly available Chinese GPT-2 \cite{zhao2019uer}.\footnote{\url{https://huggingface.co/uer/gpt2-chinese-cluecorpussmall}} Same as in Section \cref{sec:experiment}, during training, we use a batch size of 128 and truncate the training samples to a maximum length of 256. On the LCCC dataset, we train (i.e., fine-tune) the models for 40k steps. As for the DailyDialog dataset, due to its smaller dataset size, we train the models for 5k steps. For optimization, we use Adam optimizer and a learning rate of 2e-5.

For each model, we use four decoding methods, including (1) greedy search; (2) beam search (beam size of $10$); (3) nucleus sampling ($p=0.95$); and (4) contrastive search ($k=5$, $\alpha=0.6$).

\textbf{Evaluation.} We rely on human evaluation to assess the model performance. Same as in Section \cref{sec:human_evaluation_detail}, we randomly select 200 dialogue contexts from the test set and ask five annotators to evaluate the generated responses plus the reference response in three dimensions: (i) coherence, (ii) fluency; and (iii) informativeness. The scores follow a 5-point Likert scale (1, 2, 3, 4, or 5).

\begin{table*}[h]
    \small
	\centering  
	\renewcommand{\arraystretch}{1.2}
	\setlength{\tabcolsep}{6pt}
	\scalebox{0.9}{
	\begin{tabular}{cccccccc}
		\hlinewd{0.75pt}
		\multirow{2}{*}{\textbf{Model}}&\multirow{2}{*}{Method}&\multicolumn{3}{c}{LCCC}&\multicolumn{3}{c}{DailyDialog}\\
		\cmidrule(lr){3-5}
		\cmidrule(lr){6-8}
        &&Coherence&Fluency&Informativeness&Coherence&Fluency&Informativeness\\
        \hline
        Agreement&-&0.73&0.61&0.57&0.64&0.60&0.55\\
        \hline
        \multirow{4}{*}{MLE}&greedy&3.01&3.27&1.97&3.28&3.51&2.92\\
        &beam&2.60&2.90&1.55&3.16&3.43&2.78\\
        &nucleus&2.78&3.55&2.64&2.67&3.58&3.42\\
        &contrastive&3.28$^{\bigstar}$&3.84$^{\bigstar}$&3.06$^{\bigstar}$&3.27&3.41&2.82\\
        \hline
        \multirow{4}{*}{SimCTG}&greedy&3.04&3.32&2.01&3.31&3.50&2.94\\
        &beam&2.57&2.93&1.59&3.19&3.45&2.79\\
        &nucleus&2.84&3.58&2.72&2.75&3.59&3.39\\
        &contrastive&\textbf{3.32}$^{\bigstar}$&\textbf{3.96}$^{\bigstar}$&\textbf{3.13}$^{\bigstar}$&\textbf{3.73}$^{\bigstar}$&\textbf{3.85}$^{\bigstar}$&\textbf{3.46}\\
        \hline
        Human&-&3.42&3.76&3.20&4.11&3.98&3.74\\
		\hlinewd{0.75pt}
	\end{tabular}}
    \caption{Human evaluation results. ${\bigstar}$ results significantly outperforms the results of greedy search, beam search, and nucleus sampling with different models. (Sign Test with p-value < 0.05).}
    	\vspace{-1.5mm}
	\label{tb:dialogue_human_evaluation}
\end{table*}

Table \ref{tb:dialogue_human_evaluation} shows the evaluation results where the first row shows strong inter-annotator agreements as measured by Fleiss$\textprime$ kappa coefficient. On both datasets, we see that SimCTG + contrastive search significantly outperforms other methods on various metrics, suggesting that our approach is generalizable to different languages and tasks. It is worth emphasizing that, on the LCCC benchmark, SimCTG + contrastive search surprisingly outperforms the human performance on the fluency metric, while performing comparably on the coherence and informativeness metrics (Sign Test with p-value > 0.4). Moreover, even \textbf{without} contrastive training, the MLE model performs significantly better when using contrastive search. This is due to the intrinsic property of Chinese language model for which the MLE objective can already yield a representation space that displays a high level of isotropy, making contrastive search directly applicable.\footnote{We provide more in-depth analyses and several generated examples on LCCC in Appendix \ref{sec:chinese_language_analysis} and \ref{sec:lccc_case_study}, respectively.} This finding is particularly attractive as it reveals the potential applicability of contrastive search on off-the-shelf (i.e., without contrastive training) language models for certain languages such as Chinese.


\section{Further Analysis}
\label{sec:detailed_analysis_section}

\begin{figure}[t]
\centering
\begin{minipage}{.5\textwidth}
  \centering
  \includegraphics[width=0.93\linewidth]{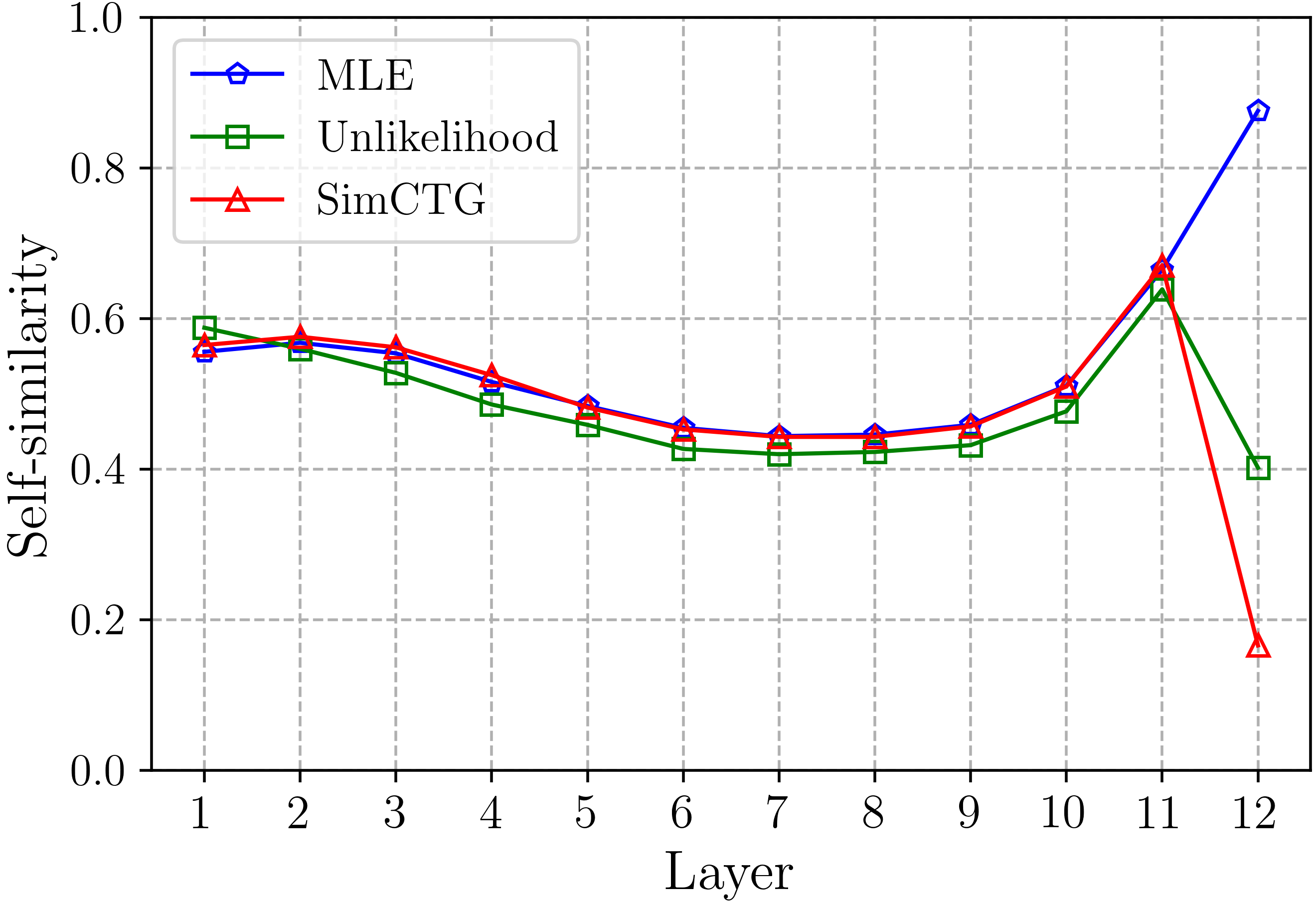}
  \captionof{figure}{Layer-wise representation self-similarity.}
  \label{fig:self_similarity}
\end{minipage}%
\begin{minipage}{.5\textwidth}
  \centering
  \includegraphics[width=0.93\linewidth]{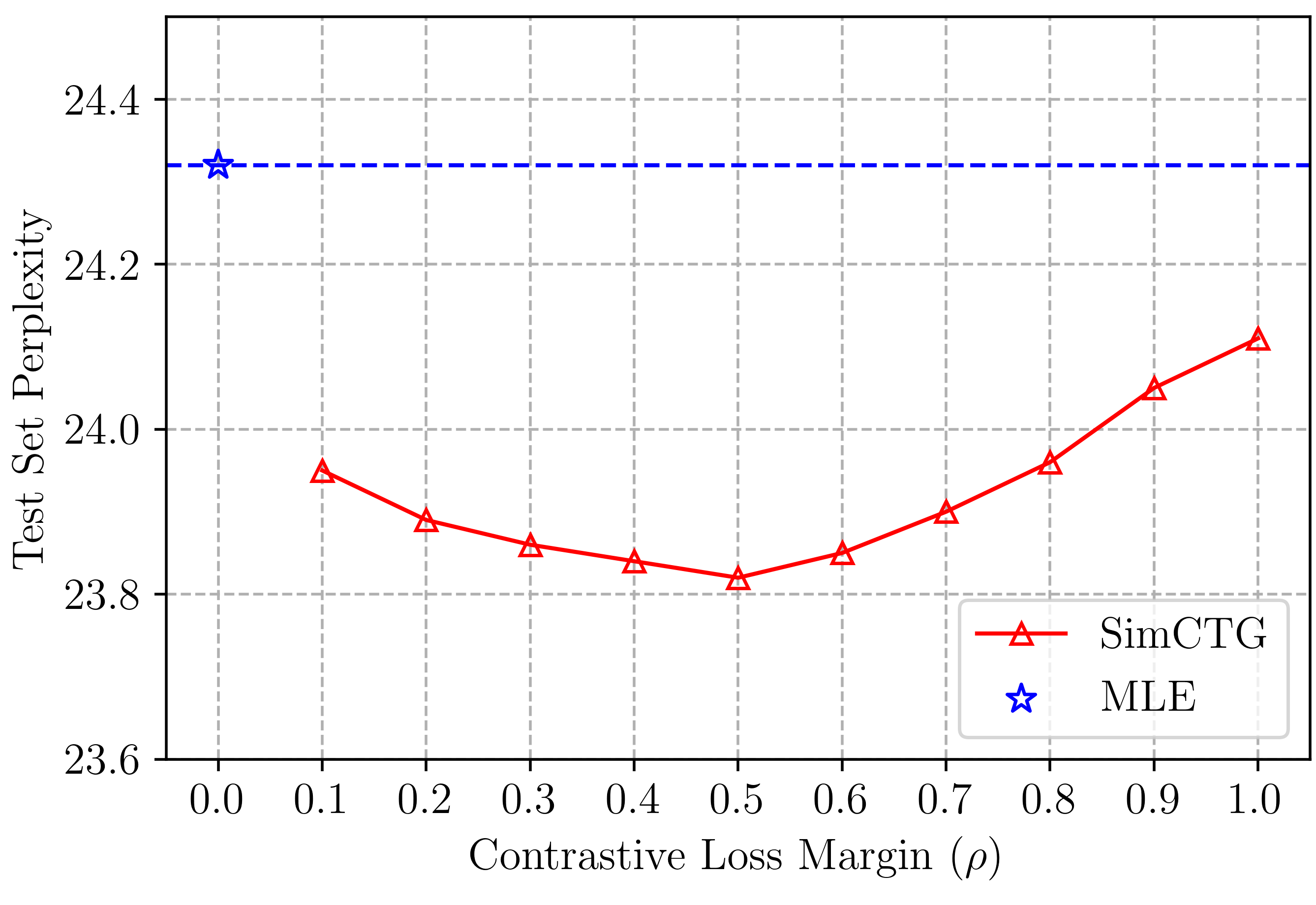}
  \captionof{figure}{The effect of contrastive margin $\rho$.}
  \label{fig:margin_vs_ppl}
\end{minipage}
\end{figure}

\subsection{Token Representation Self-similarity}
\label{sec:self_similarity}
To analyze the token representations learned by SimCTG, we follow Ethayarajh~\cite{DBLP:conf/emnlp/Ethayarajh19} and define the averaged self-similarity of token representations within a text sequence $\boldsymbol{x}$ as
\begin{equation}
\label{eq:self_similarity}
    \textup{self-similarity}(\boldsymbol{x}) = \frac{1}{|\boldsymbol{x}|\times (|\boldsymbol{x}|-1)}\sum_{i=1}^{|\boldsymbol{x}|}\sum_{j=1,j\neq i}^{|\boldsymbol{x}|}\frac{h_{x_i}^\top h_{x_j}}{\|h_{x_i}\|\cdot\|h_{x_j}\|},
\end{equation}
where $h_{x_i}$ and $h_{x_j}$ are the token representations of $x_i$ and $x_j$ produced by the model. Intuitively, a lower $\textup{self-similarity}(\boldsymbol{x})$ indicates the representations of distinct tokens within the sequence $\boldsymbol{x}$ are less similar to each other, therefore being more discriminative. 

We use texts from  Wikitext-103 test set and compute the self-similarity of token representations over different layers for different models. 
Figure \ref{fig:self_similarity} plots the results averaged over all samples. We see that, in the intermediate layers, the self-similarity of different models are relatively the same. In contrast, at the output layer (layer 12), SimCTG's self-similarity becomes notably lower than other baselines. We note that the Unlikelihood model also yields more discriminative representations than MLE, but its language model accuracy is lower than MLE and SimCTG as shown in Table \ref{tb:main_result}. On the other hand, SimCTG obtains the most discriminative and isotropic representations while maintaining the best language model accuracy, which further validates the clear advantage of our proposed approach.

\subsection{The Effect of Contrastive Loss Margin}
Next, we analyze the effect of contrastive loss margin $\rho$ (Eq. (\ref{eq:cl})). To this end, we fine-tune the GPT-2 by varying $\rho$ from $0.1$ to $1.0$ and measure the model perplexity on the Wikitext-103 test set. Figure \ref{fig:margin_vs_ppl} plots the results of different $\rho$ along with the result of the MLE baseline. Note that, when $\rho=0$, SimCTG is equivalent to MLE (Section \cref{sec:contrastive_training}). From Figure \ref{fig:margin_vs_ppl}, we see that the contrastive training always helps to improve the perplexity as compared with MLE. However, when $\rho$ is either too small (e.g., $0.1$) or large (e.g., $1.0$), the learned representation space of the model would be either less or too isotropic, leading to a sub-optimal perplexity. In our experiments, the most suitable margin $\rho=0.5$.

\subsection{Contrastive Search versus Nucleus Sampling}
Then, we provide an in-depth comparsion between our proposed contrastive search and the current state of the art, nucleus sampling. To this end, we compare the results of SimCTG using these two decoding methods. Specifically, we vary the probability $p$ for nucleus sampling and the $\alpha$ (Eq. (\ref{eq:score})) for contrastive search to generate results using prefixes from Wikitext-103 test set.\footnote{For contrastive search, we only vary the value of $\alpha$ and keep $k$ constant to $8$ as described in Section \cref{sec:experiment}. In Appendix \ref{appendix:ablation_study}, we provide detailed ablation studies on the effect of both $k$ and $\alpha$ in contrastive search.} We evaluate the results from two aspects: (1) generation diversity and (2) perplexity of the generated text (gen-ppl). Both metrics are described in Section \cref{sec:generation_quality_metric}. Figure \ref{fig:diversity_vs_gen-ppl} plots the results of different methods along with the human performance. For nucleus sampling, when $p$ is small (i.e., $p\leq 0.7$), its generation perplexity is comparable to that of human. However, the diversity is notably lower than human performance, meaning it stuck in undesirable repetition loops \cite{DBLP:conf/iclr/HoltzmanBDFC20}. On the other hand, when $p$ is large (i.e., $p\geq 0.95$), the generation diversity is close to that of human but the generation perplexity is significantly higher. Such high perplexity means the generated text is very \textit{unlikely}, therefore being low quality. As for contrastive search, when $\alpha\in[0.5,0.8]$, it yields generation diversity and perplexity that are both comparable to human performance. These results demonstrate the superiority of contrastive search as it better balances the trade-off between the generation diversity and perplexity.

\subsection{Decoding Latency Comparison}
\label{sec:inference_latency}
\begin{figure}[tb]
\centering
\begin{minipage}{.5\textwidth}
  \centering
  \includegraphics[width=0.93\linewidth]{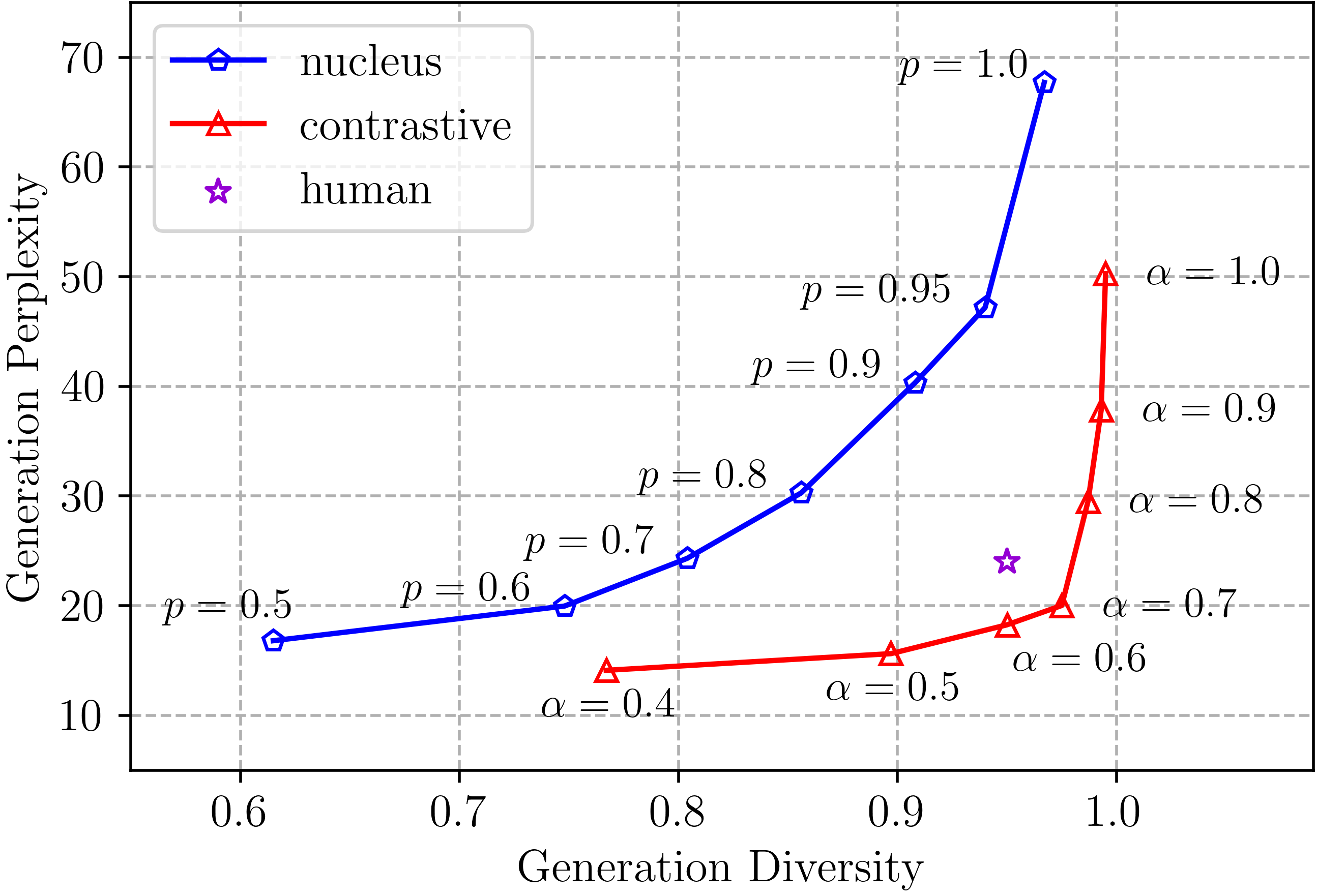}
  \captionof{figure}{Contrastive search vs nucleus sampling.}
  \label{fig:diversity_vs_gen-ppl}
\end{minipage}%
\begin{minipage}{.5\textwidth}
  \centering
  \includegraphics[width=0.93\linewidth]{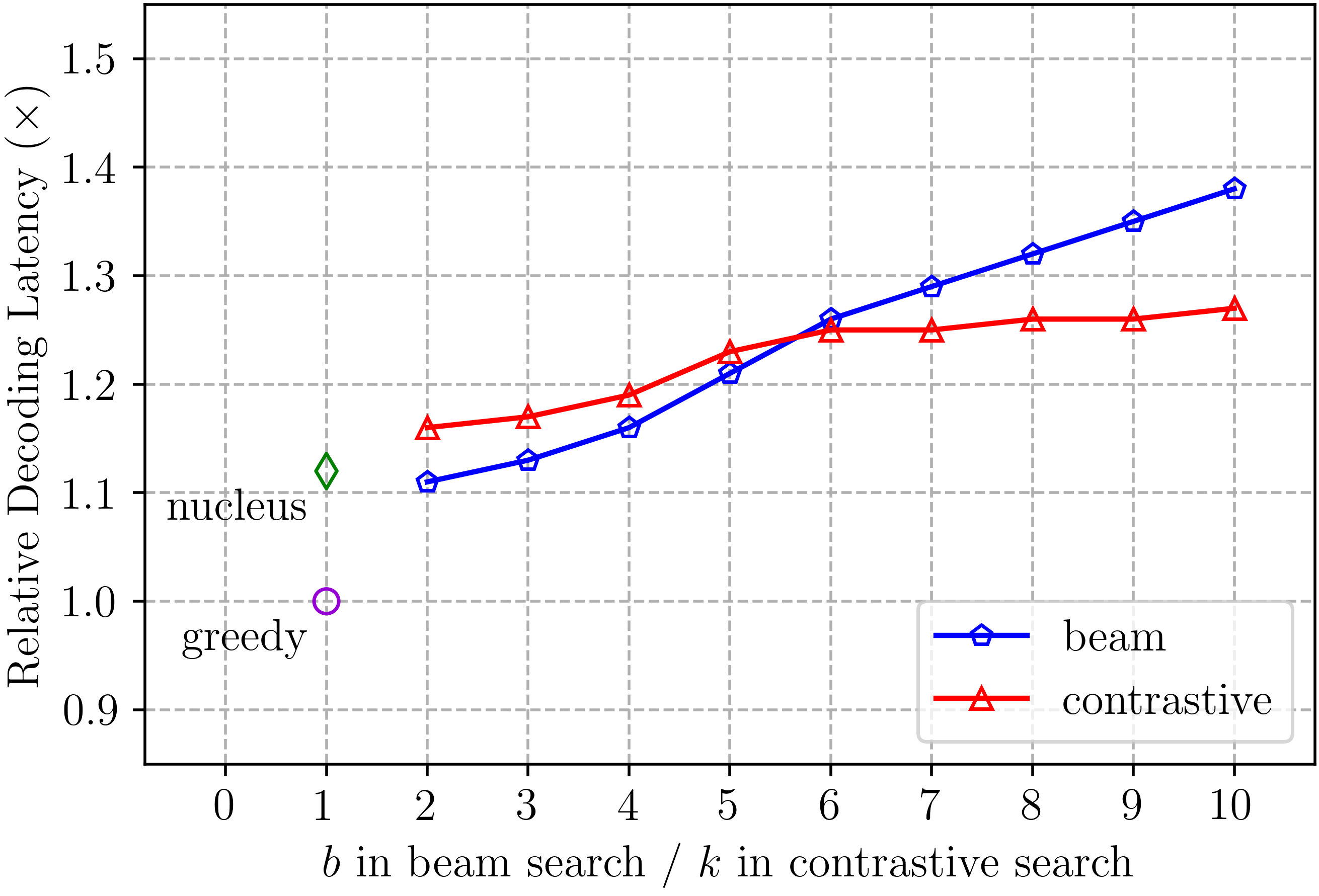}
  \captionof{figure}{Inference latency comparison.}
  \label{fig:inference_latency}
\end{minipage}
\end{figure}

We compare the decoding latency of different decoding methods using SimCTG. For beam search and contrastive search, we vary the beam width $b$ and the $k$ in Eq. (\ref{eq:score}). The latency is measured by generating fixed length text continuations on Wikitext-103 test cases with a batch size of $1$. In Figure \ref{fig:inference_latency}, we show the averaged relative decoding latency of different methods. We see that greedy search is the fastest method and the latency of different methods are generally comparable with each other. Comparing contrastive search with beam search, when $b$ and $k$ are small (i.e., $\leq6$), their latency are nearly identical. When $b$ and $k$ gets larger (i.e., $>6$), contrastive search becomes faster. In summary, these comparison results further verify the practical usage of contrastive search.

\begin{table*}[h]
    \small
	\centering  
	\renewcommand{\arraystretch}{1.2}
	\setlength{\tabcolsep}{6pt}
	\scalebox{0.745}{
	\begin{tabular}{ccccc}
		\hlinewd{0.75pt}
		\textbf{prefix}&\makecell[l]{Butt criticized Donald's controls in certain situations in the game, as well as the difficulty \\of some levels and puzzles. Buchanan also criticized the controls, calling}&div.&coh.&gen-ppl\\
		\hline
		
		beam&\makecell[l]{them " too easy " and " too difficult ". \textcolor{red}{He also criticized the game's graphics, stating that the game's}\\ \textcolor{red}{graphics were " too simplistic " and " too simplistic ".} He also criticized the game's graphics, stating\\ that the game's graphics were " too simplistic " and " too simplistic ". He also criticized the graphics,\\ stating that the game's graphics were " too simplistic " and " too simplistic ". He also criticized the\\ graphics, stating that the game's graphics were " too simplistic " and " too simplistic ". He also\\ criticized the graphics, stating that the game's graphics were" ...}&0.02&0.605&7.91\\
		\hline
		nucleus&\makecell[l]{them " offensively rudimentary ". After the game was released, Buchanan wrote a number of game\\ reviews about his review of the game, primarily summarizing what \textcolor{blue}{he thought the reviewers had}\\ \textcolor{blue}{enjoyed about it.} \textcolor{blue}{Buchanan was pleased with how consistently these reviews awarded the game.} \\For one, \textcolor{blue}{he praised the puzzles}, compared them to Nos. 1 and 2 and \textcolor{blue}{enjoyed the construction of the}\\ \textcolor{blue}{artificial intelligence in the game's forms}. \textcolor{blue}{He found no fault with the puzzles}, even in those situations\\ where they seemed repetitive. According to an Xbox Magazine review, " every time ...}&0.98&0.597&37.83\\
		\hline
		contrastive&\makecell[l]{them " unimpressive " and a " nightmare " of an experience to play with players unfamiliar with\\ \textcolor{teal}{Tetris}. On the other hand, his opinion was shared by other reviewers, and some were critical of the\\ \textcolor{teal}{game}'s technical design for the Wii version of \textcolor{teal}{Tetris}. In addition, Tintin's review included a quote\\ from Roger \textcolor{teal}{Ebert}, who said that \textcolor{teal}{Tetris} was better than the original \textcolor{teal}{game} due to its simplicity and\\ ease of play. \textcolor{teal}{Ebert}'s comments were included in the \textcolor{teal}{game}'s DVD commentary, released on March\\ 22, 2010. It is unclear if any of the video commentary was taken from ...}&0.98&0.626&19.64\\
		\hlinewd{0.75pt}
	\end{tabular}}
    \caption{\textbf{Case Study}: The beam search produces degeneration repetitions (highlighted in \textcolor{red}{red}) and the nucleus sampling produces text that has incoherent semantics with respect to the prefix (highlighted in \textcolor{blue}{blue}). The reasonable repetitions produced by contrastive search are highlighted in \textcolor{teal}{green}. The ``div.'' and ``coh.'' stand for diversity and coherence metrics. (best viewed in color)}
	\label{tb:case_study}
\end{table*}

\subsection{Case Study}
In Table \ref{tb:case_study}, we present generated examples of SimCTG with different decoding methods given a specific prefix.\footnote{We refer to Appendix \ref{appendix:case_study} for more generated examples of SimCTG.} From the results, we see that beam search produces undesirable sequence-level repetitions, resulting in low diversity and low generation perplexity. On the other hand, in the prefix, the person ``Buchanan'' \textit{criticizes} the game. However, the result from nucleus sampling displays a contradicted semantic, resulting in a low coherence score as well as a high generation perplexity. As for contrastive search, it generates a text that is semantically consistent to the prefix with a proper generation perplexity while obtaining the same diversity as that of the nucleus sampling. Additionally, it is worth emphasizing that, while the degeneration penalty in Eq. (\ref{eq:score}) encourages the model to generate diverse outputs, contrastive search is still able to generate reasonable repetitions as highlighted in Table \ref{tb:case_study}. This is due to the incorporation of model confidence in Eq. (\ref{eq:score}) which enables the model to repeat the important content (e.g., person names or entity names) from the previous context like humans do.

\begin{figure*}[h] 
  \centering    
  \setlength{\abovecaptionskip}{3pt}
  \includegraphics[width=1.0\textwidth]{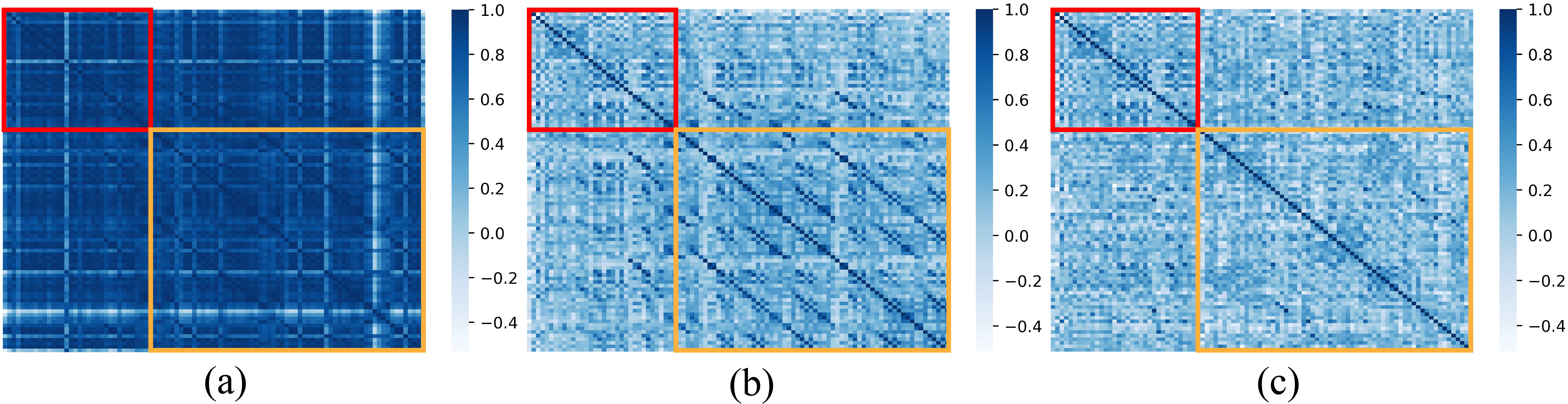}
  \caption{(a) MLE + beam search; (b) SimCTG + beam search; (c) SimCTG + contrastive search. The token similarity matrix of the prefix and the generated text are highlighted in red and yellow.}
  \label{fig:decoding_heatmap}
  \vspace{-1.5mm}
\end{figure*}

\subsection{Comparison of Token Similarity Matrix}
To better understand how contrastive search works, in Figure \ref{fig:decoding_heatmap}, we show the generated token similarity matrix of SimCTG using beam search and contrastive search. For a better comparsion, we also include the result of MLE using beam search. All results are produced with the same prefix as in Table \ref{tb:case_study}. The red and yellow boxes highlight the similarity matrix of the prefix and the generated text. Firstly, we see that, the MLE + beam search yields a very dense similarity matrix, meaning that its token representations are indiscriminative. In addition, the high similarity scores in its off-diagonal entries clearly show the degeneration repetitions. Secondly, for SimCTG + beam search, we observe a desirable similarity matrix of the prefix which is sparse and isotropic. However, degeneration repetitions still exist in the generated result as shown in Figure \ref{fig:decoding_heatmap}(b). Lastly, for SimCTG + contrastive search, the entire similarity matrix is sparse and isotropic, showing that it successfully solves the model degeneration. These observations are in line with our motivations as described in Section \cref{sec:introduction}.

\section{Conclusion}
\label{sec:conclusion}
In this work, we show that the degeneration of neural language models stems from the anisotropic nature of their token representations. We present a new approach, \textit{SimCTG}, for training the language model such that it obtains an isotropic and discriminative representation space. In addition, we introduce a novel decoding method, \textit{contrastive search}, which works coherently with the proposed SimCTG. Extensive experiments and analyses are conducted on three benchmarks from two languages. Both automatic and human evaluations demonstrate that our approach substantially reduces model degeneration and significantly outperforms current state-of-the-art text generation approaches.

\section*{Acknowledgments}
The first author would like to thank Jialu Xu and Huayang Li for their insightful discussions and supports. Many thanks to our anonymous reviewers, area chairs, and senior area chairs for their suggestions and comments.

\bibliographystyle{plain}
\bibliography{reference}

\clearpage

\section*{Checklist}
\begin{enumerate}

\item For all authors...
\begin{enumerate}
  \item Do the main claims made in the abstract and introduction accurately reflect the paper's contributions and scope?
    \answerYes{}
  \item Did you describe the limitations of your work?
    \answerYes{See Appendix~\ref{appendix:future_work}.}
  \item Did you discuss any potential negative societal impacts of your work?
    \answerNA{}
  \item Have you read the ethics review guidelines and ensured that your paper conforms to them?
    \answerYes{}
\end{enumerate}

\item If you are including theoretical results...
\begin{enumerate}
  \item Did you state the full set of assumptions of all theoretical results?
    \answerNA{}
        \item Did you include complete proofs of all theoretical results?
    \answerNA{}
\end{enumerate}

\item If you ran experiments...
\begin{enumerate}
  \item Did you include the code, data, and instructions needed to reproduce the main experimental results (either in the supplemental material or as a URL)?
    \answerYes{We provide the code and the instructions to re-implement our results as a supplementary material to this paper.}
  \item Did you specify all the training details (e.g., data splits, hyperparameters, how they were chosen)?
    \answerYes{We specify the details in Section \cref{sec:experiment} and  \cref{sec:dialogue_experiment}.}
        \item Did you report error bars (e.g., with respect to the random seed after running experiments multiple times)?
    \answerNo{We did not run multiple times for our experiments due to computational constraints.}
        \item Did you include the total amount of compute and the type of resources used (e.g., type of GPUs, internal cluster, or cloud provider)?
    \answerYes{We describe the computational details in Appendix J.}
\end{enumerate}

\item If you are using existing assets (e.g., code, data, models) or curating/releasing new assets...
\begin{enumerate}
  \item If your work uses existing assets, did you cite the creators?
    \answerYes{We cite the authors of the datasets and the code of the models in Section \cref{sec:experiment} and \cref{sec:dialogue_experiment}.}
  \item Did you mention the license of the assets?
    \answerNA{The datasets are publicly available.}
  \item Did you include any new assets either in the supplemental material or as a URL?
    \answerYes{We provide the code and the instructions to re-implement our results as a supplementary material to this paper.}
  \item Did you discuss whether and how consent was obtained from people whose data you're using/curating?
    \answerNA{The datasets are publicly available.}
  \item Did you discuss whether the data you are using/curating contains personally identifiable information or offensive content?
    \answerNo{We use the standard datasets, which are well known in literature, and there are no personally identifiable information or offensive content at the best of the community knowledge.}
\end{enumerate}

\item If you used crowdsourcing or conducted research with human subjects...
\begin{enumerate}
  \item Did you include the full text of instructions given to participants and screenshots, if applicable?
    \answerYes{We provide the human evaluation guidelines in Appendix~\ref{sec:evaluation_guideline}.}
  \item Did you describe any potential participant risks, with links to Institutional Review Board (IRB) approvals, if applicable?
    \answerNA{}
  \item Did you include the estimated hourly wage paid to participants and the total amount spent on participant compensation?
    \answerYes{We provide the details of participant compensation in Appendix~\ref{sec:evaluation_guideline}.}
\end{enumerate}

\end{enumerate}

\newpage

\appendix
\addcontentsline{toc}{section}{Appendix} 
\part{Appendix} 
\parttoc 
\clearpage

\section{Future Work}
\label{appendix:future_work}
For future work, we would like to suggest three research directions based on our study.
\begin{itemize}
    \item Our proposed contrastive loss $\mathcal{L}_{\textup{CL}}$ in Eq. (\ref{eq:cl}) is designed to treat all other tokens within the same sequence as negative samples. However, we do acknowledge that there might be a suitably small fraction of tokens (within the same sequence) that share similar semantic meanings even with different surface forms. We believe the current formulation of the contrastive loss might be further improved by taking this aspect into consideration and we leave it to our future work. 
    \item One limitation of the proposed contrastive search is that it is a deterministic decoding method. It would be interesting and useful to incorporate a certain level of stochasticity into the decoding process. One plausible approach is to combine contrastive search with stochastic sampling methods. For instance, given the prefix, we could first generate a few tokens (e.g., 1$\sim$3 tokens) with nucleus sampling. Then, we switch to contrastive search for the remaining steps. In Appendix \ref{sec:diverse_contrastive_search}, we provide some preliminary experiment results on incorporating stochasticity into contrastive search.
    \item Our approach is architecture agnostic and can be applied to any generation model. Future research could focus on adapting it to other tasks than open-ended text generation (i.e., constrained text generation), such as machine translation and document summarization.
\end{itemize}

\section{Related Work}
\textbf{Neural Text Generation} is a core component in many NLP applications. It can be generally categorized into two classes (1) constrained generation; and (2) open-ended generation. 

Constrained generation tasks are always defined over a set of (input, output) pairs, where the output is a transformation of the input following specific constrains. Some typical examples include machine translation \cite{DBLP:conf/nips/SutskeverVL14,DBLP:journals/corr/BahdanauCB14,DBLP:conf/emnlp/LuongPM15}, text summarization \cite{DBLP:conf/acl/SeeLM17,DBLP:conf/eacl/SuCWVBLC21}, and data-to-text generation \cite{DBLP:conf/emnlp/WisemanSR17,DBLP:conf/emnlp/SuVWFC21,DBLP:conf/emnlp/SuMBC21,DBLP:journals/corr/abs-2201-05966}. As the output is tightly scoped by the input, the generation of repetition and unnaturalness are not that problematic, therefore maximization-based decoding methods such as beam search generally perform well. Still, different variants of beam search have been explored to further improve the model performance in constrained generation tasks \cite{DBLP:conf/acl/KleinKDSR17,DBLP:conf/icml/KoolHW19,DBLP:conf/naacl/LuWZBBC21,DBLP:journals/corr/abs-2112-08726}.

Open-ended text generation, on the other hand, imposes less constrain on the generated text. It aims at producing text that is natural, coherent and informative with respect to the human-written prefix (i.e., context). Several typical applications include story generation \cite{DBLP:conf/acl/LewisDF18,su2022language}, contextual text completion \cite{radford2019language}, and dialogue systems \cite{DBLP:journals/taslp/SuWCBKC21,DBLP:journals/corr/abs-2109-14739}. However, due to the challenges posed by the increased level of freedom, conventional maximization-based decoding methods (e.g., greedy and beam search) often produce undesirable repetition and unnaturalness in the generated text. To alleviate model degeneration, different sampling approaches \cite{DBLP:conf/acl/LewisDF18,DBLP:conf/iclr/HoltzmanBDFC20,meister2022typical} have been proposed to generate text by drawing samples from less likely vocabularies. Welleck \emph{et al.}~\cite{DBLP:conf/iclr/WelleckKRDCW20} tackled model degeneration from another perspective by introducing unlikelihood objective into the training of the language model. 

\textbf{Contrastive Learning.} Generally, contrastive learning methods aim to teach the model to distinguish observed data points from fictitious negative samples. They have been widely applied to various research areas. In the field of computer vision, contrastive learning has been shown to benefit tasks like image \cite{DBLP:journals/corr/abs-1807-03748} and video \cite{DBLP:conf/icra/SermanetLCHJSLB18} representation learning.  Chen \emph{et al.}~\cite{DBLP:conf/icml/ChenK0H20} proposed a simple framework, SimCLR, for learning contrastive visual representations. Recently, Radford \emph{et al.}~\cite{DBLP:conf/icml/RadfordKHRGASAM21} and Jia \emph{et al.}~\cite{DBLP:conf/icml/JiaYXCPPLSLD21} applied contrastive learning for the pre-training of language-image models.

In the field of NLP, contrastive learning has recently gained much more attention. Numerous contrastive approaches have been proposed to learn better token-level \cite{DBLP:journals/corr/abs-2111-04198}, sentence-level \cite{DBLP:journals/corr/abs-2102-08473,DBLP:conf/emnlp/0001VKC21,DBLP:conf/emnlp/GaoYC21}, and discourse-level \cite{DBLP:conf/acl/Su0ZLBC0C020,DBLP:journals/corr/abs-2110-06612,an2022cont,krishna2022rankgen} representations. Beyond representation learning, contrastive learning has also been applied to other NLP applications, such as name entity recognition (NER) \cite{DBLP:journals/corr/abs-2109-07589}, document summarization \cite{DBLP:conf/acl/Liu020a}, and knowledge probing for pre-trained language models \cite{DBLP:journals/corr/abs-2110-08173}. 

Our work, to the best of our knowledge, is the first effort on applying contrastive learning to address neural text degeneration. We hope our findings could facilitate future research in this area.

\section{Software Package}
In this section, we illustrate the use of the accompanying Python package, available on Github\footnote{\url{https://github.com/yxuansu/SimCTG/tree/main/simctg}} and installable via pip\footnote{\url{https://pypi.org/project/simctg/}} as \texttt{pip install simctg -{}-upgrade}.

Below, we show how to replicate our result in Table~\ref{tb:case_study} with our provided package. More details can be found in our open-sourced repository\footnote{\url{https://github.com/yxuansu/SimCTG}}.

\begin{lstlisting}[language=Python,caption=Example usage of the SimCTG package,captionpos=b]
import torch
# load the language model
from simctg.simctggpt import SimCTGGPT
model_name = r'cambridgeltl/simctg_wikitext103'
model = SimCTGGPT(model_name)
model.eval()
tokenizer = model.tokenizer

# prepare input
prefix_text = # The prefix text in Table 4
print ('Prefix is: {}'.format(prefix_text))
tokens = tokenizer.tokenize(prefix_text)
input_ids = tokenizer.convert_tokens_to_ids(tokens)
input_ids = torch.LongTensor(input_ids).view(1,-1)

# generate result with contrastive search
beam_width, alpha, decoding_len = 8, 0.6, 128
output = model.fast_contrastive_search(input_ids=input_ids, 
                        beam_width=beam_width, alpha=alpha,
                        decoding_len=decoding_len) 
print("Output:\n" + 100 * '-')
print(tokenizer.decode(output))

\end{lstlisting} %

\begin{table*}[h]
    \small
	\centering  
	\renewcommand{\arraystretch}{1.2}
	\setlength{\tabcolsep}{6pt}
	\scalebox{0.76}{
	\begin{tabular}{ccccccccccc}
		\hlinewd{0.75pt}
		\textbf{Model}&Size&Objective&ppl$\downarrow$&acc$\uparrow$&conicity$\downarrow$&self-similarity$\downarrow$&\textbf{Method}&diversity$\uparrow$&MAUVE$\uparrow$&coherence$\uparrow$\\
        \hline
        \multirow{4}{*}{Transformers}&\multirow{4}{*}{117M}&\multirow{2}{*}{MLE}&\multirow{2}{*}{26.60}&\multirow{2}{*}{35.62}&\multirow{2}{*}{0.50}&\multirow{2}{*}{0.22}&nucleus&0.89&0.81&0.541\\
        &&&&&&&contrastive&0.90&0.83&0.561\\
        \cline{3-11}
        &&\multirow{2}{*}{SimCTG}&\multirow{2}{*}{\textbf{26.55}}&\multirow{2}{*}{\textbf{36.03}}&\multirow{2}{*}{\textbf{0.47}}&\multirow{2}{*}{\textbf{0.19}}&nucleus&0.89&0.82&0.543\\
        &&&&&&&contrastive&\textbf{0.91}&\textbf{0.85}&\textbf{0.566}\\
        \hline
        \multirow{4}{*}{GPT-2-small}&\multirow{4}{*}{117M}&\multirow{2}{*}{MLE}&\multirow{2}{*}{24.32}&\multirow{2}{*}{39.63}&\multirow{2}{*}{0.90}&\multirow{2}{*}{0.86}&nucleus&0.94&0.90&0.577\\
        &&&&&&&contrastive&0.24&0.18&0.599\\
        \cline{3-11}
        &&\multirow{2}{*}{SimCTG}&\multirow{2}{*}{\textbf{23.82}}&\multirow{2}{*}{\textbf{40.91}}&\multirow{2}{*}{\textbf{0.43}}&\multirow{2}{*}{\textbf{0.18}}&nucleus&0.94&0.92&0.584\\
        &&&&&&&contrastive&\textbf{0.95}&\textbf{0.94}&\textbf{0.610}\\
        \hline
        \multirow{4}{*}{GPT-2-large}&\multirow{4}{*}{774M}&\multirow{2}{*}{MLE}&\multirow{2}{*}{16.57}&\multirow{2}{*}{43.34}&\multirow{2}{*}{0.46}&\multirow{2}{*}{0.20}&nucleus&0.94&0.91&0.583\\
        &&&&&&&contrastive&\textbf{0.95}&\textbf{0.96}&0.623\\
        \cline{3-11}
        &&\multirow{2}{*}{SimCTG}&\multirow{2}{*}{\textbf{16.53}}&\multirow{2}{*}{\textbf{43.47}}&\multirow{2}{*}{\textbf{0.42}}&\multirow{2}{*}{\textbf{0.17}}&nucleus&\textbf{0.95}&0.93&0.591\\
        &&&&&&&contrastive&\textbf{0.95}&\textbf{0.96}&\textbf{0.626}\\
        \hline
        Human&-&-&-&-&-&-&-&0.95&1.00&0.644\\
		\hlinewd{0.75pt}
	\end{tabular}}
    \caption{Experimental results of different language models on Wikitext-103. $\uparrow$ means higher is better and $\downarrow$ means lower is better. The results of GPT-2-small are copied from Table~\ref{tb:main_result}.}
    	\vspace{-1.5mm}
	\label{tb:further_experiments}
\end{table*}

\section{Experiments on Different Language Models}
\label{appendix:different_model_experiment}

In this section, we further test the generalization ability of our approach with different language models on the Wikitext-103 benchmark. In addition to the GPT-2-small model (i.e. 12 Transformer layers with 12 attention heads) that we consider in Section~\cref{sec:experiment}, we include (i) a vanilla Transformers (i.e. without any pre-training) with the same parameter size as GPT-2-small; and (ii) a larger pre-trained model, GPT-2-large, that consists of 36 Transformer layers with 20 attention heads. The training of different language models follows the same procedure as described in Section~\cref{sec:experiment}. To measure the isotropy of the language model, we include the conicity metric~\cite{sharma2018towards} as well as the self-similarity metric (Eq.~(\ref{eq:self_similarity})). A lower conicity or self-similarity indicates the representation space of the language model better follows an isotropic distribution.

Table~\ref{tb:further_experiments} presents the experimental results. We observe that our approach (i.e. SimCTG + contrastive search) performs the best on all evaluated models, suggesting the clear generalization ability of our approach. Another interesting finding is that, for vanilla Transformers and GPT-2-large, the model trained with MLE naturally displays a high level of isotropy. A similar phenomenon is also observed in language models from other languages, such as Chinese (see Appendix~\ref{sec:chinese_language_analysis}). In such cases, our proposed contrastive search can be directly applied and yields superior performances. This further points out the huge potential of contrastive search in other much larger and stronger language models such as GPT-3~\cite{DBLP:conf/nips/BrownMRSKDNSSAA20} and OPT~\cite{zhang2022opt}. We leave the rigorous investigation on the isotropic properties of different language models to our future work.

\section{Ablation Study on the Hyperparameters of Contrastive Search }
\label{appendix:ablation_study}
\begin{figure*}[h] 
  \centering    
  \setlength{\abovecaptionskip}{3pt}
  \includegraphics[width=0.6\textwidth]{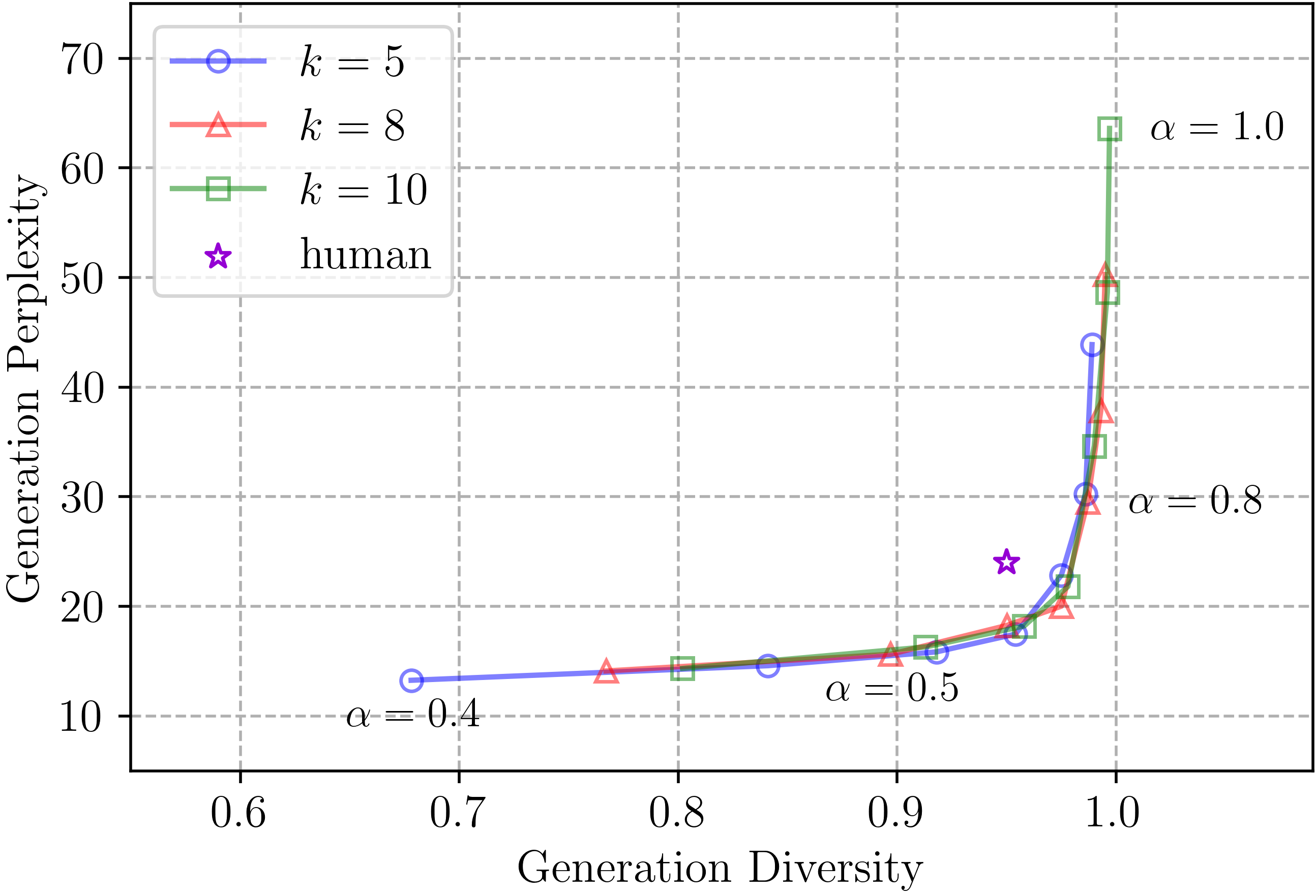}
  \caption{Ablation study on the hyperparameters of contrastive search.}
  \label{fig:hyperparameter_analysis}
  \vspace{-1.5mm}
\end{figure*}

Here, we present a detailed ablation study on the hyperparameters (i.e., $k$ and $\alpha$ in Eq. (\ref{eq:score})) of contrastive search. Specifically, we simultaneously vary the value of $k$ and $\alpha$. $k$ is chosen from $\{5, 8, 10\}$ and $\alpha$ is chosen from  $\{0.4, 0.5, 0.6, 0.7, 0.8, 0.9, 1.0\}$. For evaluation, we report the generation diversity and generation perplexity on the test set of Wikitext-103. The results are plotted in Figure \ref{fig:hyperparameter_analysis}.  We see that, when $k$ is constant, the increase of $\alpha$ generally increases the generation diversity and generation perplexity. When $\alpha$ is constant, a larger $k$ also leads to the increased generation diversity as well as generation perplexity. Nonetheless, for different $k$, the overall trends are relatively the same and the value of $\alpha$ has more impact on the generated results. In practice, our recommended selection range of $k$ and $\alpha$ are $k\in [5,10]$ and $\alpha\in[0.5,0.8]$, as these settings produce results that are more similar to  human-written texts as judged by generation diversity and generation perplexity.

\section{Gen-ppl Results Measured by Different Models}
\label{appendix:gen_ppl}

\begin{table*}[h]
\small
\parbox{.49\linewidth}{
\centering
	\renewcommand{\arraystretch}{1.2}
	\setlength{\tabcolsep}{6pt}
	\scalebox{0.8}{
	\begin{tabular}{cccccc}
		\hlinewd{0.75pt}
		&greedy&beam&nucleus&contrastive&human\\
		\hline
		MLE&7.77&6.48&48.82&9.43&\multirow{3}{*}{24.86}\\
		Unlike.&39.02&37.38&76.22&46.03&\\
		SimCTG&8.01&6.87&47.64&\textbf{20.53}&\\
		\hlinewd{0.75pt}
	\end{tabular}
	}
\caption{The results of gen-ppl measured by the model trained with MLE.}
\label{tb:mle_gen_ppl_results}
}
\hfill
\parbox{.49\linewidth}{
\centering
	\renewcommand{\arraystretch}{1.2}
	\setlength{\tabcolsep}{6pt}
	\scalebox{0.8}{
	\begin{tabular}{cccccc}
		\hlinewd{0.75pt}
		&greedy&beam&nucleus&contrastive&human\\
		\hline
		MLE&13.18&11.67&58.01&15.94&\multirow{3}{*}{29.62}\\
		Unlike.&44.13&42.67&71.13&47.82&\\
		SimCTG&12.34&10.98&55.24&\textbf{23.47}&\\
		\hlinewd{0.75pt}
	\end{tabular}
	}
\caption{The results of gen-ppl measured by the model trained with Unlikelihood.}
\label{tb:unlikelihood_gen_ppl_results}
}
\end{table*}

In Table \ref{tb:mle_gen_ppl_results} and \ref{tb:unlikelihood_gen_ppl_results}, we show the gen-ppl (detailed in \cref{sec:generation_quality_metric}) results of different methods as measured by the model trained with MLE and Unlikelihood, respectively. As we use different models to measure gen-ppl, the results in Table \ref{tb:mle_gen_ppl_results} and \ref{tb:unlikelihood_gen_ppl_results} are slightly different from the ones in Table \ref{tb:main_result}. Nontheless, we can draw the same conclusion as in Section \cref{sec:automatic_evaluation_result} that SimCTG + contrastive search is the best performing method as it obtains the generation perplexity that is closest to the human-written text.

\section{Human Evaluation Guidelines}
\label{sec:evaluation_guideline}
Given the human-written prefix, please evaluate the system's result with respect to the following features: (1) Coherence; (2) Fluency; and (3) Informativeness. In the following, we provide some guidelines regarding how to judge the quality of the system's result in terms of different features.

\subsection{Coherence}
This metric measures whether the system's result is semantically and factually consistent with the human-written prefix. The definitions of different scores are:
\begin{itemize}
    \item \textbf{[5]}: The system's result is perfectly in line with the semantic meaning defined by the prefix. And all its content is factually supported by or can be logically inferred from the prefix.
    \item \textbf{[4]}: The system's result is very related to the prefix but with some minor errors that does not affect its overall relevance with respect to the prefix.
    \item \textbf{[3]}: The system's result is, to some extent, relevant to the prefix with some errors that display minor semantic inconsistency or contradiction. 
    \item \textbf{[2]}: At the first glance, the system's result seems to be related to the prefix. But with careful inspection, the semantic inconsistency can be easily spotted.
    \item \textbf{[1]}: The system's result is obviously off-the-topic or it is semantically contradicted to the content contained in the prefix. 
\end{itemize}

\subsection{Fluency}
This metric measures the fluency of the system's result. The definitions of different scores are:
\begin{itemize}
    \item \textbf{[5]}: The system's result is human-like, grammatically correct, and very easy to understand.
    \item \textbf{[4]}: Choose this score when you are hesitant between the score 3 and score 5.
    \item \textbf{[3]}: The system's result contains minor errors but they do not affect your understanding.
    \item \textbf{[2]}: Choose this score when you are hesitant between the score 1 and score 3.
    \item \textbf{[1]}: The system's result does not make sense and it is unreadable.
\end{itemize}

\subsection{Informativeness}
This metric measures the diversity, informativeness, and interestingness of the system's result. The definitions of different scores are:
\begin{itemize}
    \item \textbf{[5]}: The system's result is very informative and contains novel content. In addition, it displays a high level of diversity and it is enjoyable to read.
    \item \textbf{[4]}: Choose this score when you are hesitant between the score 3 and score 5.
    \item \textbf{[3]}: The system's result contains some new information and it displays a certain level of diversity.
    \item \textbf{[2]}: Choose this score when you are hesitant between the score 1 and score 3.
    \item \textbf{[1]}: The system's result is dull, repetitive, and does not have new information. All its content has already been provided in the prefix.
\end{itemize}

\textbf{Participant Compensation.} In each experiment (i.e., open-ended text generation and open-domain dialogue generation), we hire 5 annotators to conduct the human evaluation. For every task, each annotator is paid by \$400.

\begin{figure*}[h] 
  \centering    
  \setlength{\abovecaptionskip}{3pt}
  \includegraphics[width=0.65\textwidth]{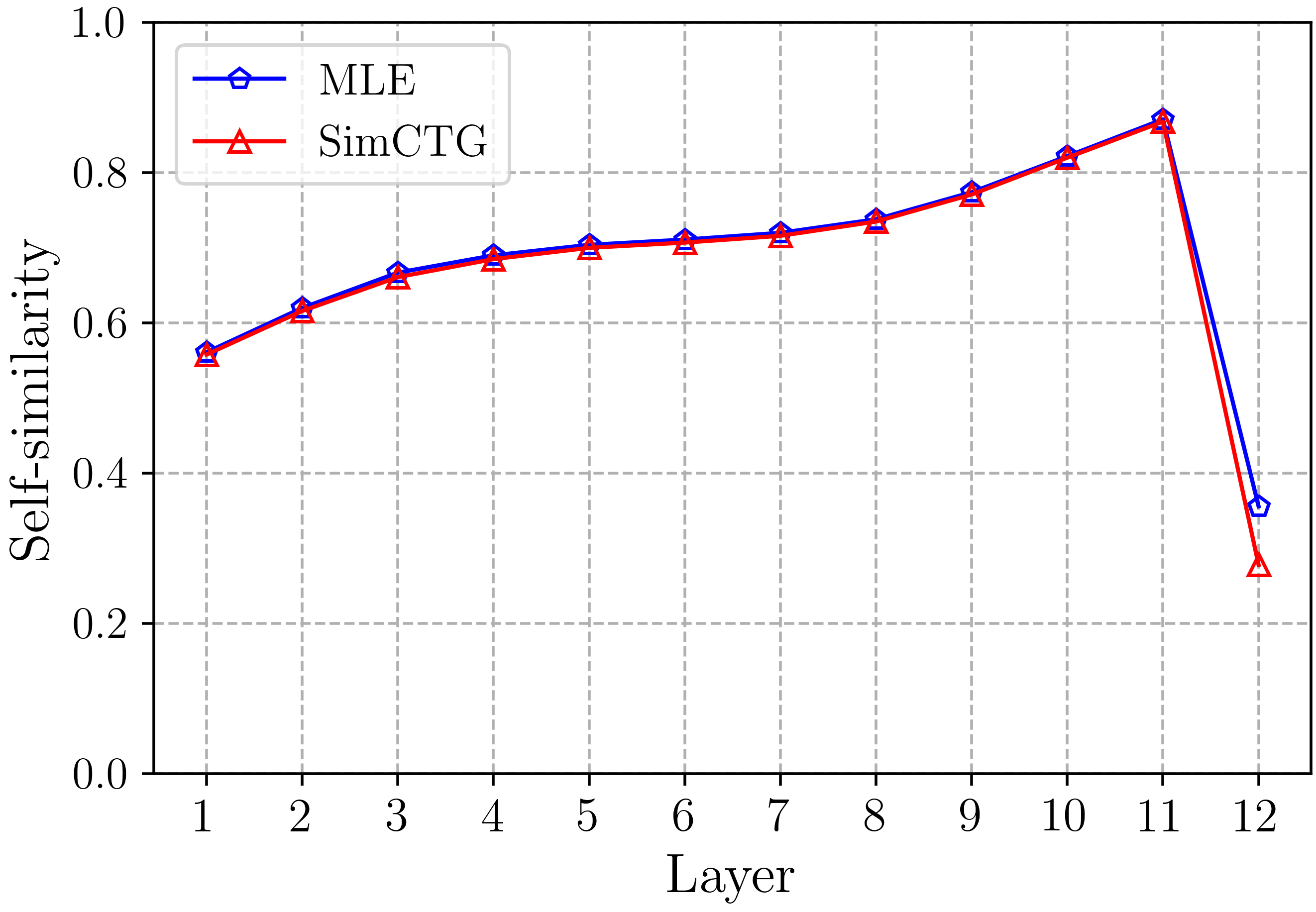}
  \caption{Layer-wise self-similarity of Chinese language models.}
  \label{fig:chinese_self_similarity}
  \vspace{-1.5mm}
\end{figure*}

\section{Self-similarity of Chinese Language Models}
\label{sec:chinese_language_analysis}
We follow the same procedure as described in Section \cref{sec:self_similarity} to measure the token self-similarity of Chinese language models. Specifically, we use the test set of LCCC benchmark and compute the model's self-similarity. Figure \ref{fig:chinese_self_similarity} plots the layer-wise token self-similarity of the MLE and SimCTG models. We see that in all layers (including the final layer), the MLE model displays a similar self-similarity with respect to SimCTG. This observation is quite different from what we see from English language models as shown in Figure \ref{fig:self_similarity}, where the self-similarities of SimCTG and MLE are notably different in the final layer. We conjecture that this discrepancy might come from the intrinsic property of different languages. For English, current state-of-the-art methods always represent the text into subword units, such as BPE \cite{DBLP:conf/acl/SennrichHB16a}, and the same subword could be over-shared by many different contexts. Thus, the representations of distinct subwords become less distinguishable which naturally leads to the anisotropy in their representations.\footnote{However, we should also note that, for larger English models (e.g., GPT-2-large), this conjecture not longer holds as demonstrated in Appendix~\ref{appendix:different_model_experiment}. This urges us to conduct more thorough investigations on the isotropic properties of language models across different sizes as well as different languages. We will leave these investigations to our future work.} On the other hand, languages like Chinese are naturally represented by basic units, i.e., characters. Such natural unit boundary of text alleviates the over-sharing of characters in different contexts. As a result, even the vanilla MLE objective can obtain a representation space that displays a high level of isotropy. 

This isotropic property of Chinese language model is particularly attractive as contrastive search can be directly applied even \textbf{without} contrastive training as shown in Table \ref{tb:dialogue_human_evaluation}. In addition, we expect contrastive search could be used on off-the-shelf language models that are trained with MLE in other languages whose texts are naturally tokenized by characters (e.g., Korean and Japanese). This remains to be rigorously tested in our future work.

\begin{table*}[h]
    \small
	\centering  
	\renewcommand{\arraystretch}{1.2}
	\setlength{\tabcolsep}{6pt}
	\scalebox{1.0}{
	\begin{tabular}{cccc}
		\hlinewd{0.75pt}
		&\textbf{MLE}&\textbf{Unlikelihood}&\textbf{SimCTG}\\
		\hline
		Train FLOPs&8.08e16&8.91e16&8.20e16\\
		Parameters&117M&117M&117M\\
		\hlinewd{0.75pt}
	\end{tabular}}
    \caption{Training efficiency comparison.}
    	\vspace{-1.5mm}
	\label{tb:train_flops}
\end{table*}

\section{Training Efficiency Comparison}
\label{appendix:training_resource}
In this part, we compare the training efficiency of different methods (i.e., MLE, Unlikelihood, and SimCTG). To this end, we compute the total floating point operations (FLOPs) required for the training of different models on Wikitext-103. The details of training setup are provided in Section \cref{sec:experiment}. Table \ref{tb:train_flops} shows the results, from which we see that SimCTG is more efficient than the unlikelihood method. Comparing with MLE, SimCTG only introduces an negligible 1.48\% extra computational overhead, which further verifies the practical usage of SimCTG.

\section{Generated Examples on Open-domain Dialogue Generation}
\label{sec:lccc_case_study}

In Table \ref{tb:chinese_dialogue_example}, we show some generated responses of our approach (i.e., SimCTG + contrastive search) plus the reference response on examples from the test set of the Chinese LCCC benchmark. We see that, given the dialogue context, our approach is able to generate responses that are both grammatically fluent and semantically consistent with the dialogue context. These results further demonstrate the generality of our approach across different languages and tasks.

\begin{table*}[h] 
	\centering    
	\setlength{\abovecaptionskip}{3pt}
	\includegraphics[width=0.95\textwidth]{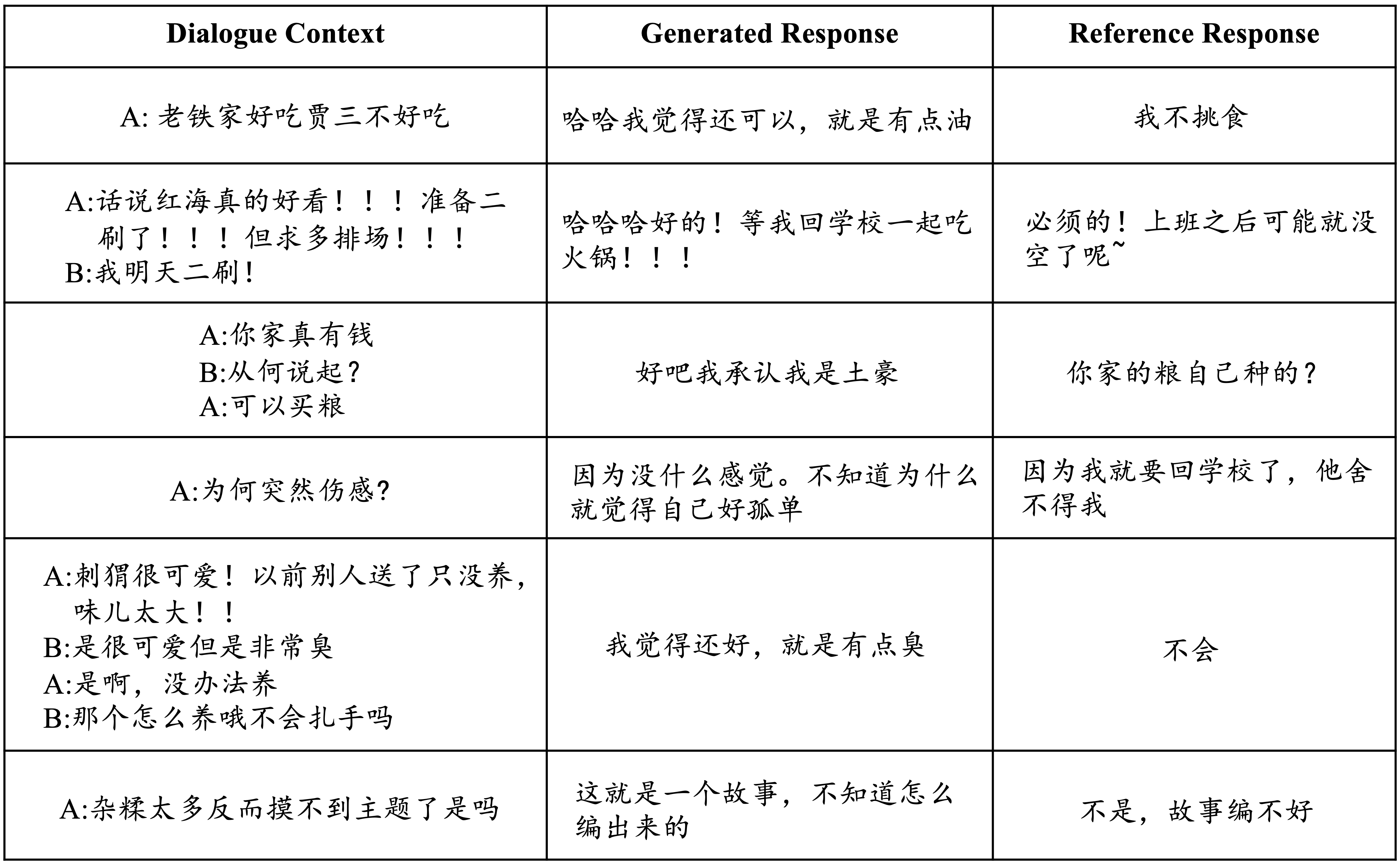}
	\caption{Case study on the LCCC dataset.}
	\label{tb:chinese_dialogue_example}
\end{table*}

\section{More Generated Examples of SimCTG + Contrastive Search}
\label{appendix:case_study}
In Table \ref{tb:more_examples}, we provide more generated examples of SimCTG + contrastive search based on prefixes from Wikitext-103. The details of the decoding procedure are described in Section \cref{sec:experiment}.

\begin{table*}[t]
    \small
	\centering  
	\renewcommand{\arraystretch}{1.4}
	\setlength{\tabcolsep}{6pt}
	\scalebox{0.95}{
	\begin{tabular}{cc}
		\hlinewd{0.75pt}
		\textbf{prefix}&\makecell[l]{The Olmec colossal heads are at least seventeen monumental stone representations of human \\heads sculpted from large basalt boulders. The heads date from at least before}\\
		\hline
		\textbf{result}&\makecell[l]{the Roman-era, and have been interpreted as representing a group of men and women that lived\\ in the city of Arundel, the same site on which Alexander the Great's tomb was built. In addition,\\ there is evidence to suggest that the city was part of the empire of the Persians and their conquests\\ in Asia Minor during the 7th century BC. This may have been an intentional move to avoid confusion\\ with the Roman city of Acre, which is located in the foothills of the mountain range near the Roman\\ fort of Caesarea. Other evidence points to the construction ...}\\
		\hline
		\hline
		\textbf{prefix}&\makecell[l]{Although Rifenburg finished fourth among midwestern Heisman voters in 1948, he did not finish\\ among the top eight. By comparison, Notre Dame end Leon Hart won}\\
		\hline
		\textbf{result}&\makecell[l]{the Heisman Trophy at the end of the 1948 season after a 4 – 0 victory over Michigan on the final\\ day of the regular season. Hart was named to the Big Ten All-Pac-10 first team, and his Heisman\\ Trophy was awarded to Notre Dame's Bobby Bowden as the team's all-time leader in passing yards\\ ( 1,000 ) and completions ( 2,065 ). The game ended in a 42 – 21 victory for Notre Dame, which led\\ the nation in total offense ( 4,000 yards ) and touchdowns ...}\\
		\hline
		\hline
		\textbf{prefix}&\makecell[l]{Kujira developed from a broad area of disturbed weather as a tropical depression on April 9 well\\ removed from any landmasses. Shortly after development, Kuj}\\
		\hline
		\textbf{result}&\makecell[l]{ira became extratropical, which would have caused the storm to turn to the west-northwest in the\\ Gulf of Mexico on April 10. However, due to the proximity of the cyclone to an area of low\\ pressure, Kujira's winds were less than 10 mph ( 15 km / h ) at the time of landfall. By April 11,\\ Hurricane Hunters estimated that there were 5,000 people in the area, of whom 1,000 were located\\ in Puerto Rico. The storm weakened to a tropical storm early on April 12, and dissipated the next\\ day. Later that day ...}\\
		\hline
		\hline
		\textbf{prefix}&\makecell[l]{The island is composed of a variety of volcanic and sedimentary rock formations from the Devonian\\ period. At that time the Scottish landmass formed part of the Old}\\
		\hline
		\textbf{result}&\makecell[l]{North Sea, and it was inhabited by an extensive and diverse population of sea-dwelling birds that had\\ lived in the islands for many centuries. There is evidence of human habitation in the Isles of Scilly\\ and St Albans, and some archaeological evidence is in the form of stone mounds dating to the 12th\\ century. The most northerly of these is the Murchison Formation, which is located south of the island's\\ mouth and extends to the northwest as far as Aberdeenshire. Other sites include the Isle of Man, Isle of\\ Wight, Isle of Orkney, The Loch Ness Monster ...}\\
		\hline
		\hline
		\textbf{prefix}&\makecell[l]{A report published in the American magazine Time a fortnight later described the incident as " a \\genuine horror story, calculated to make the most alarming of Rhodesian doomsday}\\
		\hline
		\textbf{result}&\makecell[l]{scenarios ". In an interview with BBC Radio 4's Tom Pertwee, he said : " There's no evidence that\\ anyone in Rhodesia, other than the police, had done anything to provoke it ". On 20 April 2013, the\\ government announced that it had found no evidence of involvement by Rhodesia in the attack, and\\ would continue its investigation as soon as possible. The Independent Commission Against Corruption\\ ( ICAO ) was appointed to investigate the incident in December 2013, and is the first body to do so in the\\ UK. ICAO Director-General Mark Rowley said it was " the most ...}\\
		\hline
		\hline
		\textbf{prefix}&\makecell[l]{The American poet Ezra Pound was introduced to the group in April 1909 and found that their ideas were\\ close to his own. In particular, Pound's studies of}\\
		\hline
		\textbf{result}&\makecell[l]{German Expressionism ( 1891 – 1903 ) helped him develop a critical attitude towards the subject, and he\\ had the opportunity to speak at the first meeting of the group on 25 April 1909, at New York's\\ Waldorf-Astoria Hotel. This was the first time Pound had been invited to speak at a German Expressionist\\ conference, and it proved to be a success. Pound became an admirer of Expressionism from the outside,\\ and was able to attract many of its members by providing them with information and writing, such as letters\\ to German Expressionists in Germany, as well as photographs of their works ...}\\
		\hline
		\hline
		\textbf{prefix}&\makecell[l]{The Olmec civilization developed in the lowlands of southeastern Mexico between 1500 and 400 BC. The\\ Olmec heartland lies on the Gulf Coast of Mexico}\\
		\hline
		\textbf{result}&\makecell[l]{, about 1,500 miles ( 2,500 km ) west of Acapulco County, where there is a population of 1,500 people. This\\ is the largest city in the state, and it is home to the city's most famous tourist attraction, the El Centro de las\\ Amazons, which has been named by Mexican historian José Martínez Sánchez for its beauty and architecture.\\ The city, along with its surroundings, were important in the development of the Maya civilization, and in later\\ Mesoamerica and Latin America as well. Today, Mexico's ...}\\
		\hline
		\hline
		\textbf{prefix}&\makecell[l]{The discovery of a colossal head at Tres Zapotes in the nineteenth century spurred the first archaeological\\ investigations of Olmec culture by Matthew Stirling in 1938.}\\
		\hline
		\textbf{result}&\makecell[l]{He showed that the remains of Olmec were in good condition, and suggested that there was something to be\\ said for the fact that they had lived long enough to be recognized as an important part of the culture and its\\ history. Stirling's work was the first to establish what was to become known as the Peruvian-Spanish language,\\ or Peru : the language of the people of Peru and Central America. In doing so, he drew on archaeological\\ evidence and other sources to support his theory of the cultural origins of the Peruvian people. This is reflected\\ in the name " Peruvian ", which stands for Peruvian ...}\\
		\hlinewd{0.75pt}
	\end{tabular}}
    \caption{More generated examples of SimCTG + contrastive search on Wikitext-103.}
	\label{tb:more_examples}
\end{table*}

\clearpage

\section{Diverse Contrastive Search}
\label{sec:diverse_contrastive_search}
In this part, we present a stochastic version of contrastive search (i.e., diverse contrastive search) which is described in Appendix~\ref{appendix:future_work}. Specifically, given the prefix with length of 32, we first generate \textbf{2} tokens using nucleus sampling with $p=0.95$, then we use contrastive search to generate the remaining 126 tokens (i.e., 128 generated tokens in total). 

\begin{table*}[h]
    \small
	\centering  
	\renewcommand{\arraystretch}{1.3}
	\setlength{\tabcolsep}{6pt}
	\scalebox{0.9}{
	\begin{tabular}{cc}
		\hlinewd{0.75pt}
		\textbf{prefix}&\makecell[l]{Butt criticized Donald's controls in certain situations in the game, as well as the difficulty of some levels and\\ puzzles. Buchanan also criticized the controls, calling}\\
		\hline
		&\multicolumn{1}{c}{\textit{Contrastive Search}}\\
		\hline
		\textbf{result}&\makecell[l]{them " unimpressive " and a " nightmare " of an experience to play with players unfamiliar with Tetris. On\\ the other hand, his opinion was shared by other reviewers, and some were critical of the game's technical\\ design for the Wii version of Tetris. In addition, Tintin's review included a quote from Roger Ebert, who said\\ that Tetris was better than the original game due to its simplicity and ease of play. Ebert's comments were\\ included in the game's DVD commentary, released on March 22, 2010. It is unclear if any of the video\\ commentary was taken from ...}\\
		\hline
		&\multicolumn{1}{c}{\textit{Diverse Contrastive Search}}\\
		\hline
		\textbf{[1]}&\makecell[l]{the game " unplayable " and adding that, " It's a game in which you have to be careful not to fall into\\ the trap of thinking too much about your opponents' strengths and weaknesses. " On the other hand,\\ his review of the game, published in the New York Times, found little to complain about in comparison\\ to some of the other games in the series. In particular, The Escapist's Brian O 'Toole wrote that the game\\ was an " absolute joy ", and " one of the best-selling games of all time ". O 'Toole concluded by saying\\ that although ...}\\
		\hline
		\textbf{[2]}&\makecell[l]{it " a complete waste of time " and " unplayable ". On the other hand, his review of Baldur's Gate II was\\ positive, with Buchanan commenting that, " Baldur's Gate II is an adventure game in its own right, full of\\ fun and challenge that makes you want to go back to the first game in your life. " Buchanan felt that there\\ were too many elements in the game for players to enjoy without some level-playing to be enjoyable at the\\ same time. He concluded by saying that Baldur's Gate II's controls were well-balanced, and that players ...}\\
		\hline
		\textbf{[3]}&\makecell[l]{the choice of " a simple jump button to perform a 'jump-and-a-bop' or more complex 'jump-and-a-bop' "\\ an error and a waste of time. On the other hand, Tintin was critical of the game's design, writing that there\\ was " too much going on " at the beginning of the game, and " not enough time " in the final cutscene for the\\ player to make it through the game at all. He felt that the gameplay was lacking in some areas, such as the ...}\\
		\hlinewd{0.75pt}
	\end{tabular}}
    \caption{Generated results of SimCTG with diverse contrastive search.}
	\label{tb:diverse_contrastive_search}
\end{table*}

Table \ref{tb:diverse_contrastive_search} shows three generated results with diverse contrastive search using the same prefix as in Table \ref{tb:case_study}. We see that only sampling \textbf{2} tokens at the start is enough to produce a diverse set of results. In future work, we will investigate other more sophisticated extensions of contrastive search.

\end{document}